\documentclass[preprint,12pt]{elsarticle}
\usepackage{amssymb}
\usepackage{multicol}
\usepackage{color}
\usepackage{xspace}
\usepackage{enumerate}
\usepackage{graphicx}
\usepackage{algorithm}
\usepackage{algorithmicx}
\usepackage{algpseudocode}
\usepackage{multirow}
\usepackage{epstopdf}
\usepackage{booktabs}
\usepackage{color}
\usepackage{footmisc}
\usepackage{float}
\floatname{algorithm}{Algorithm}
\newcommand{\tabincell}[3]{\begin{tabular}{@{}#1@{}}#2\end{tabular}}

\usepackage{amsmath} 

\usepackage{amssymb}

\UseRawInputEncoding
\begin{document}

\begin{frontmatter}



\title{\textbf{Representation Evaluation Block-based Teacher-Student Network for the Industrial Quality-relevant Performance Modeling and Monitoring}}

\author[Ecust]{Dan Yang}
\author[Ecust]{Xin Peng\corref{mycorrespondingauthor}}

\ead{xinpeng@ecust.edu.cn}
\author[Ecust]{Yusheng Lu}
\author[Ecust]{Haojie Huang}
\author[Ecust,Tongji]{Weimin Zhong\corref{mycorrespondingauthor}}
\cortext[mycorrespondingauthor]{Corresponding author}
\ead{wmzhong@ecust.edu.cn}

\address[Ecust]{Key Laboratory of Advanced Control and Optimization for Chemical Processes, Ministry of Education, East China University of Science and Technology, Shanghai, 200237, China}

\address[Tongji]{Shanghai Institute of Intelligent Science and Technology, Tongji University. Shanghai 200092, China}

\begin{abstract}
Quality-relevant fault detection plays an important role in industrial processes, while the current quality-related fault detection methods based on neural networks main concentrate on process-relevant variables and ignore quality-relevant variables, which restrict the application of process monitoring. Therefore, in this paper, a fault detection scheme based on the improved teacher-student network is proposed for quality-relevant fault detection. In the traditional teacher-student network, as the features differences between the teacher network and the student network will cause performance degradation on the student network, representation evaluation block (REB) is proposed to quantify the features differences between the teacher and the student networks, and uncertainty modeling is used to add this difference in modeling process, which are beneficial to reduce the features differences and improve the performance of the student network. Accordingly, REB and uncertainty modeling is applied in the teacher-student network named as uncertainty modeling teacher-student uncertainty autoencoder (TSUAE). Then, the proposed TSUAE is applied to process monitoring, which can effectively detect faults in the process-relevant subspace and quality-relevant subspace simultaneously. The proposed TSUAE-based fault detection method is verified in two simulation experiments illustrating that it has satisfactory fault detection performance compared to other fault detection methods.

\end{abstract}


\end{frontmatter}

\footnote{The authors declare no competing financial interest. This work is supported by National Natural Science Foundation of China (61890930-3,61925305 and Basic Science Center Program: 61988101), Fundamental Research Funds for the Central Universities, 222201917006, Shanghai Sailing Program under Grant 18YF1405200.}
\section{Introduction}
Casualties and property damages may happen if severe accidents occur to the industrial processes, which have drawn attention to ensure the practical processes operate in normal conditions. Process monitoring plays an important role in terms of preventing accidents, which can be classified as model-based methods, knowledge-based methods and data-driven methods \cite{Ge2017, Li2019, Liu2021}. As the large amount of data are generated in the actual process, the data-driven method does not need to fully understand the mechanism of the system and can quickly build models, so the data-driven method has been widely concerned in the fault detection \cite{Luo2020, Huang2020}. Typical data-driven methods such as principal component analysis (PCA) \cite{Hu2016}, partial least squares (PLS) \cite{Tong2019, NikzadLangerodi2020}, independent component analysis (ICA) \cite{Peng2017, Spurek2018} and robust bayesian network \cite{Chen2020} have been widely used in fault detection. 

The above methods are main used to monitor the fluctuation and the abnormal conditions in the process subspace, but in the actual industrial, more attentions may be paid to whether the failure caused by process-relevant variables will lead to the change of the product quality. However, due to the high cost and the difficulty of quality-relevant variables measurement, the usual way is to explore the relationship between the easily measured process-relevant variables and the difficultly measured quality-relevant variables, so as to monitor the fluctuation of quality-relevant variables through the change of process-relevant variable \cite{Qin2003, Park2020}. Quality-relevant fault detection based on PLS and its improved algorithm are the most commonly used methods \cite{Yin2016}. Zhou et al. proposed a total projection to latent structures (T-PLS) algorithm, which improves the prediction performance by separating the orthogonal and correlated subspace, and removing data that is not helpful to the prediction of the quality-relevant variables \cite{Zhou2009}. Li et al. proposed a  combined index based on T-PLS and failure reconstruction methods \cite{Li2010}. Peng et al. proposed the total kernel projection to the latent structures (t-KPLS) for the nonlinear phenomenon in quality monitoring \cite{Peng2013}. Zhao et al. proposed multi-space generalization of total projection to latent structures (MST-PLS) to deal with the multiplicity of process-relevant space, which improves the performance of real-time monitoring \cite{Zhao2013}. Although the above data-driven algorithms have achieved good performance in the fault detection of industrial processes, it is difficult to effectively deal with data with the strong nonlinear relationship due to its shallow structure.

The neural network (NN) can fully mine the relationship between data and variables by using multi-layer nonlinear mapping, so as to realize the deep features extraction. Using NN for fault detection can better overcome the lack of learning ability of traditional algorithms and make better use of data information in industrial processes. Adaptive network \cite{Jin2020, Zhu2020}, back propagation network \cite{Han2020a, Ren2014} and radial basis network \cite{Qays2020, Li2019a} are typical networks used in fault detection. However, these methods tend to focus on process-relevant variables or quality-relevant variables, and separate the features extraction of quality-relevant variables from the features extraction of process-relevant variables, which will reduce the performance and generalization ability of monitoring. Therefore, we consider using the teacher-student(T-S) network, which contains two networks, to extract the features of process-relevant variables and represent the features of quality-relevant variables simultaneously \cite{Yan2019}. In the T-S network, the teacher network is used to extract the features of process-relevant variables, and the student network is used to predict the combination of process-relevant variables and quality-relevant variables \cite{Hinton2015, Lecun2015}. The traditional T-S network pre-trains the teacher network and then trains the student network under the supervision of the features extracted by the teacher network, which is beneficial to the student network. In this way, the student network  can be regarded as the knowledge distillation of teacher network. However, the mimic student features must be different from the teacher feature, which degradates the performance of the student network \cite{Ba2014}. 

Thus, we proposed teacher-student uncertainty autoencoder (TSUAE) to reduce the features differences between the teacher network and the student network and improve the performance of the student network. In the TSUAE, representation evaluation block (REB) is used to evaluate the features differences between the teacher network and the student network. In REB, the evaluation of the testing phase is the student features, and the evaluation of the training phase is the teacher features with the feedback of the differences between the teacher network and the student network. To describe the feedback of the differences, a hyperparameter denoting the negative feedback rate is set. The negative feedback is used to alleviate the performance degradation caused by the differences. However, the performance degradation caused by differences can not be eliminated by negative feedback. Inspired by uncertainty autoencoder (UAE) that the noise is added in the training of autoencoder \cite{Grover2019}, the uncertainty model is considered to estimate the differences and added the differences in modeling processs. In this way, the performance degradation is reduced through modeling the differences rather than be alleviated by the negative feedback. Accordingly, a method of fault detection based on TSUAE is proposed to extract the features of process-relevant variables and quality-relevant variables simultaneously through two networks. After training the teacher-student network, representative features from the student network can be available through REB, and then the combination of the evaluated process-relevant variables and quality-relevant variables are calculated. The errors of the process-relevant variables and the quality-relevant variables are calculated and used to estimate the threshold in a normal state through kernel density estimation (KDE) \cite{Zhou2018}. After that, we can judge whether the statistics of a new sample are under the threshold, and realize the process monitoring according to the process variables.

The main contributions of this paper are as follows: 

(1) To improve the performance of the student network in the T-S network, REB is proposed to evaluate the features differences between teacher network and student network.

(2) Uncertainty modeling is introduced to estimate the differences through REB for reducing the impact caused by differences during modeling. Accordingly, TSUAE, a network based on T-S framework, is proposed.

(3) The training method of the T-S network is redesigned by asynchronous iteration. By this way, the differences between the teacher network and the student network in network training can be suppressed.

The rest of the paper is organized as follows: in Section 2, we introduce the T-S network and the UAE in detail, which are the preliminaries of our method. In Section 3, we first propose REB to evaluate the differences, then propose uncertainty modeling on REB as TSUAE, and finally propose a fault detection method based on TSUAE. Then, a numerical example and benchmark simulation model no.1 (BSM1) simulation are used to verify the reliability of TSUAE for fault detection in Section 4. Finally, conclusions are presented in Section 5.

\section{Preliminaries for the proposed method}

The proposed TSUAE is based on the T-S network, in which the differences between the teacher network and the student network are given back to the teacher network through uncertainty modeling. The idea of uncertainty modeling into the teacher network is inspired by UAE. In this section, T-S network and UAE are introduced in detil.

\subsection{The framework of the teacher-student network}

\begin{figure}[t]
	\vskip 0in
	\begin{center}
		\centerline{\includegraphics[width=\columnwidth]{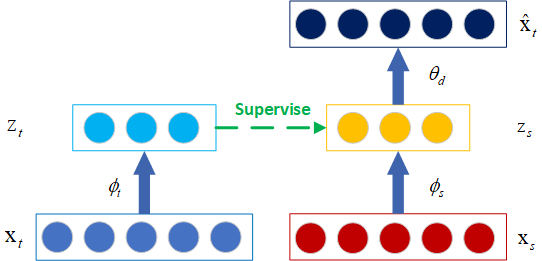}}
		\caption{The framework of the teacher-student structure}
		\label{fig1}
	\end{center}
	\vskip -0.4in
\end{figure}

In the T-S network, the student network can achieve the similar performance as by mimicking the teacher network \cite{Ba2014}.	The T-S network first uses supervised learning to train a deep teacher network, then uses unsupervised learning to train a student network with the guide of the teacher network, which is shown in Fig. \ref{fig1}. So that the prediction performance of student network would be similar to that of teacher network. The network structure of teachers and students can be formulated as:
\begin{equation}
\begin{array}{l}
\label{E1}
\mathbf{z}_{t}=f_{t}\left(\mathbf{x}_{t} ; \phi_{t}\right) \\
\mathbf{z}_{s}=f_{s}\left(\mathbf{x}_{s} ; \phi_{s}\right) \\
\hat{\mathbf{x}}_{t}=f_{d}\left(\mathbf{z_s} ; \theta_{d}\right)
\end{array}
\end{equation}

\noindent where $\mathbf{x}_{t}=\left[x_{1}, x_{2}, \ldots, x_{m}\right]^{T} \in \mathbf{R}^{m}$ is the input variable to the teacher metwork, $\mathbf{x}_{s}=\left[x_{1}, x_{2}, \ldots, x_{p}\right]^{T} \in \mathbf{R}^{p}$ is the input variable to the student metwork, $\mathbf{z}_{t} \in \mathbf{R}^{v}$ and $\mathbf{z}_{s} \in \mathbf{R}^{v}$ are the features of the networks. $\theta_{d}$ is the parameters of the decoding process, while $\phi_{t}$ and $\phi_{s}$ are the parameters of the encoding process. $f(\cdot)$ is manually set linear or nonlinear function and $\widehat{\mathbf{x}}_{t}$ is the estimate of $\mathbf{x}_{t}$.

\subsection{Uncertainty autoencoder}

To give the feedback of the features differences between two networks to the teacher network, uncertainty modeling is considered in training the teacher network, which is inspired by UAE. UAE, designed in Ref. \cite{Grover2019}, is a modification of autoencoder, which takes into account the noise in the training process of autoencoder. UAE models uncertainty in the training process by adding noise in the latent features of the encoder.

\begin{figure}[t]
	\vskip 0in
	\begin{center}
		\centerline{\includegraphics[width=\columnwidth]{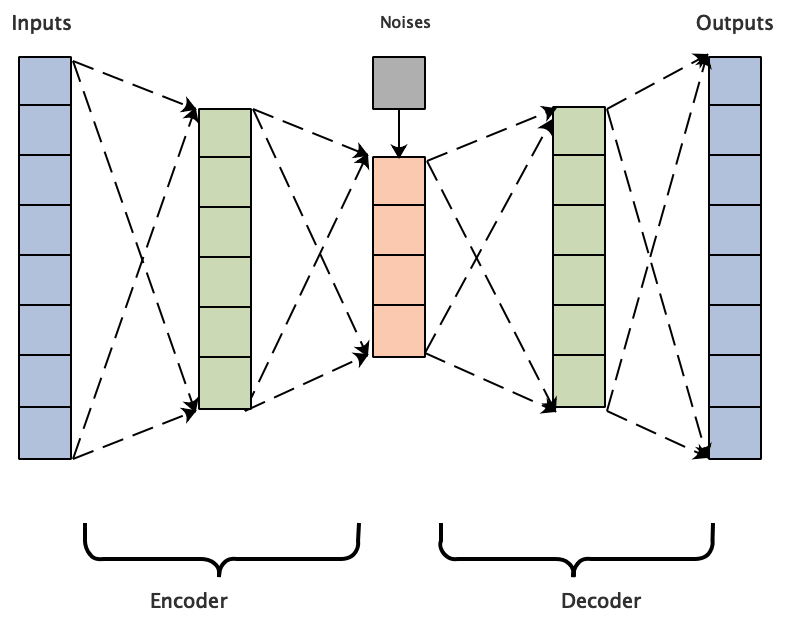}}
		\caption{The model architecture of UAE}
		\label{fig2}
	\end{center}
	\vskip -0.4in
\end{figure}

As shown in Fig. \ref{fig2}, UAE consists of two parts, i.e. the encoder and decoder. The encoder represents the edcoding process with noise through a parameterized encoding function $f$ from high dimensional signals ${{\mathbf{x}}_{t}}$ to measured data $\mathbf{z}$ , where ${{\mathbf{x}}_{t}}\in {{\mathbf{R}}^{n}}$, $\mathbf{z}\in {{\mathbf{R}}^{m}}$, $f:{{\mathbf{R}}^{n}}\to {{\mathbf{R}}^{m}}$. The encoder can be expressed according:

\begin{equation}
\label{E2}
\mathbf{z}=f({{\mathbf{x}}_{t}};\phi )+\varepsilon,
\end{equation}

\noindent where $\varepsilon$ represents external noise during the measurement and $\mathbf{z}$ is also called the representation features. Decoder corresponds to a mapping function $g$ from the measurement data $\mathbf{z}$ to the recovery data ${{\mathbf{\hat{x}}}_{t}}$, where $g:{{\mathbf{R}}^{m}}\to {{\mathbf{R}}^{n}}$:

\begin{equation}
\label{E3}
{{\mathbf{\hat{x}}}_{t}}=g(\mathbf{z};\theta ).
\end{equation}

The objective function of the autoencoder is to minimize the difference between the output of the decoder and the input of the encoder, that is, to maximize the posterior probability of the signal ${{\mathbf{x}}_{t}}$ recovered from the measured data $\mathbf{z}$:

\begin{equation}
\label{E4}
{{\max }_{\phi \in \Phi }} q_{\phi}({{\mathbf{x}}_{t}} \mid \mathbf{z}),
\end{equation}
\noindent where $q_{\phi}({{\mathbf{x}}_{t}} \mid \mathbf{z})$ is the posterior probability of the signal ${{\mathbf{x}}_{t}}$ recovered from the measured data $\mathbf{z}$.

To learn the optimal recovery $\phi \in \Phi $ with fixed uncertainty, the objective function is to maximize the logarithmic posterior probability of recovering the signal ${{\mathbf{x}}_{t}}$ from the measured data $\mathbf{z}$, which is calculated through:

\begin{equation}
\label{E5}
{{\phi }^{*}}=\arg {{\max }_{\phi \in \Phi }}\mathbb{E}_{q_{\phi}(\mathbf{x_t, z)}}\left[q_{\phi}({{\mathbf{x}}_{t}} \mid \mathbf{z})\right].
\end{equation}

\noindent where $\mathbb{E}(\cdot)$ is its expected value, ${{q}_{\phi}}({{\mathbf{{x}}}_{t}},\mathbf{{z}})$ is the joint distribution of the inputs signal ${{\mathbf{x}}_{t}}$ and the measurement data $\mathbf{z}$. As the mutual information on the measurement data and the recovery data can be expressed through:

\begin{equation}
\begin{aligned}
\label{E6}
{{I}_{\phi }}(\mathbf{{{x}_{t}}},\mathbf{z})&=-{{H}_{\phi }}({{\mathbf{x}}_{t}} \mid \mathbf{z})\\
&={\mathbb{E}_{{{q}_{\phi }}(\mathbf{{{x}_{t}},z)}}}\left[ \log {{q}_{\phi }}({{\mathbf{x}}_{t}}\mid \mathbf{z})\right]+{H}(\mathbf{{{x}_{t}}})
\end{aligned}.
\end{equation}

\noindent where ${H}(\mathbf{{{x}_{t}}})$ is the entropy of data, and serves as a constant. Thus, the objection function Eq. \ref{E5} is equivalent to maximizing the mutual information on the measurement data and the recovery data as Eq. \ref{E7}. 
\begin{equation}
\label{E7}
{{\phi }^{*}}=\arg {{\max }_{\phi \in \Phi }}\mathbb{E}_{q_{\phi}\mathbf{(x_t, z)}}\left[\log q_{\phi}({{\mathbf{x}}_{t}} \mid \mathbf{z})\right] =\arg {{\max }_{\phi \in \Phi }}{{I}_{\phi }}\mathbf{({{x}_{t}},z)}.
\end{equation}

However, estimating mutual information between high dimensional random variables can be challenging, \cite{Grover2019} gets lower bound the mutual information by introducing a variational approximation ${{p}_{\theta }}({{\mathbf{x}}_{t}}\mid\mathbf{z})$ to the model posterior $q_{\phi}\mathbf{(x_t \mid z)}$, and when the bound is tight, ${{p}_{\theta }}({{\mathbf{x}}_{t}}\mid\mathbf{z})$ is euqal to  $q_{\phi}\mathbf{(x_t \mid z)}$. Thus, the objective of UAE is given by :

\begin{equation}
\begin{aligned}
\label{E8}
\quad \max \mathcal{L}(\phi ,\theta )&=\max (\mathbb{E}_{q_{\phi}\mathbf{(x_t, z)}}\left[\log p_{\theta}({{\mathbf{x}}_{t}} \mid \mathbf{z})\right]) \\ 
\end{aligned}.
\end{equation}

\section{Teacher-student uncertainty autoencoder and its application to fault detection}

To reduce the performance degradation from the mimic student model, REB is proposed to evaluated the features differences between the teacher network and the student network, which feed back to the training process of the teacher network through uncertainty modeling. Therefore, in this section, we first introduced the REB. After that, the REB based on uncertainty modeling is proposed as TSUAE. Then, we proposed a fault detection method based on TSUAE.

\subsection{Representation evaluation block}

Though the student network intends to mimic the features from the teacher network, the features of the teacher and student network must be different, which will degrade the performance of the student network. Thus the optimal pre-trained model will suffer from unavoidable performance degradation when the student network is substituted for the teacher network. On the contrary, if we can accurately measure the differences as the feedback and consider the feedback in training, the model would be able to show the robustness and ensure the performance when using the mimic feature from the student network to substitute the features of the teacher network.

Therefore, REB is proposed to generate the virtual robust features in the training phase and the mimic student features in the testing phase according to the usage pattern. In the testing phase, the representative features of the student network are used as the imitate feature of the teacher network. In the training phase, robust features that can improve robustness of a system are generated, which represent the features of the teacher network and reflect the feedback of the features differences between two networks. REB is designed as:

\begin{equation}
\label{E9}
\text{REB}({\mathbf{z}_{t}}, {\mathbf{z}_{s}}, phase)=\left\{ \begin{aligned}
& {\mathbf{z}_{s}}\hspace*{1.3cm}(\textit{phase}='testing') \\ 
& {\mathbf{z}_{t}}+\mathbf{d}_f\hspace*{0.5cm}(\textit{phase}='training') \\ 
\end{aligned} \right.,
\end{equation}

\noindent where ${\mathbf{z}_{t}}\in {{\mathbf{R}}^{v}}$ and ${\mathbf{z}_{s}}\in {{\mathbf{R}}^{v}}$ represent the representative features of the teacher network and the student network, respectively. $\mathbf{d}_f$ is the feedback of features differences added into the training phase of the system. When REB is used in the testing phase to get the representative features of combination variables through the student network, the output of REB is considered as the representative features of the student network. When REB is used in the training phase to get the robust features that are given from the inconsistency of the two networks. Therefore, REB is a novel design to improve the robustness of the system. Specifically, if $\mathbf{d}_f$ is zero and thus virtual robust features are irrelevant with the teacher features, the teacher network is pre-trained, which is a special case of REB.

\begin{figure}[]
	\vskip 0in
	\begin{center}
		\centerline{\includegraphics[width=\columnwidth]{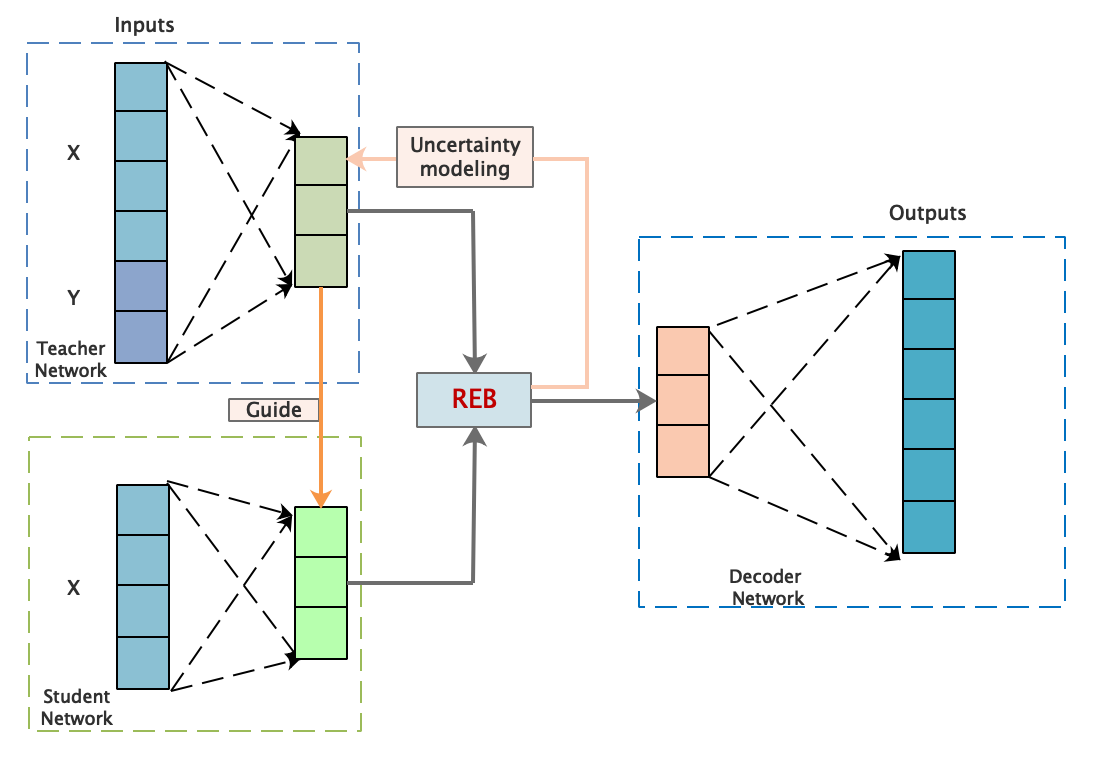}}
		\caption{The framework of the teacher-student network with representation evaluation block}
		\label{fig3}
	\end{center}
	\vskip -0.4in
\end{figure}

\subsection{Teacher-student uncertainty autoencoder based on uncrtainty modeling of REB}

To measure the feedback of features differences added in the training phase of the system, uncertainty modeling is applied in REB, in which the feedback of features differences is isotropic Gaussian noise whose volume is the maximum likelihood estimation of the features differences. Therefore, $\mathbf{d}_f \sim \mathcal{N}\left( 0,{{\sigma }^{2}}\mathbf{I} \right)$ serves as the inconsistency of the features of the two networks, whose volume is reflected from ${{\sigma }^{2}}$ and estimated through maximum likelihood estimation.

\begin{equation}
\begin{array}{l}
\label{E10}
\mathbf{d}_{f} \sim \mathcal{N}\left(0, \sigma^{2} \mathbf{I}\right) \\
\text {s.t.} \quad \mathbb{E}_{Q_{\phi s, \phi_{t}}\left(Z_{t}, Z_{s}\right)}\left[\sigma^{2}\right]=\mathbb{E}_{Q_{\phi s, \phi_{t}}\left(Z_{t}, Z_{s}\right)}\left[\left\|\mathbf{z}_{t}-\mathbf{z}_{s}\right\|_{2}^{2}\right],\\
\qquad  \quad\sigma^{2}=\mathbb{E}\left[\left\|\mathbf{z}_{t}-\mathbf{z}_{s}\right\|_{2}^{2}\right]
\end{array}
\end{equation}

\noindent where ${\mathbf{z}_{t}} \sim Z_t$, ${\mathbf{z}_{s}} \sim Z_s$, $Z_t$ and $Z_s$ are the distribution of the $\mathbf{z}_{t}$ and $\mathbf{z}_{s}$. The robust training of the teacher network is realized through the robust features generated from REB. In uncertainty modeling, the robust features are composed of teacher features and noise sampled from the uncertainty model, which endows the robustness of the training of the teacher-student network.

Uncertainty modeling on REB with the teacher-student network is shown in Fig. \ref{fig3}, which extracts representative features from inputs and realizes the robust training of the teacher network. The whole model can be described as follows:

\begin{equation}
\begin{aligned}
\label{E11}
&{\mathbf{z}_{t}}={{f}_{t}}({\mathbf{x}_{t}};{\phi }_{t}) \\ 
&{\mathbf{z}_{s}}={{f}_{s}}({\mathbf{x}};{\phi }_{s})\\ 
&\mathbf{z}=\text{REB}({\mathbf{z}_{t}},{\mathbf{z}_{s}},phase)\\ 
&{{{\hat{\mathbf{x}}}}_{t}}={{f}_{d}}(\mathbf{z};{\theta }_{d})
\end{aligned} \hspace*{0.3cm}   ,
\end{equation}

\noindent where ${\mathbf{x}_{t}}=[\mathbf{x},\mathbf{y}]$ is the combination variables $\mathbf{x}$ and variables $\mathbf{y}$. Meanwhile, ${\mathbf{x}_{t}} \sim X_t$, $X_t$ is the distribution of $\mathbf{x_t}$. $\mathbf{x}={{[{{x}_{1}},{{x}_{2}},\cdots {{x}_{m}}]}^{T}}$; $\mathbf{y}={{[{{y}_{1}},{{y}_{2}},\cdots {{y}_{p}}]}^{T}}$; $\mathbf{z}={{[{{z}_{1}},{{z}_{2}},\cdots {{z}_{v}}]}^{T}}$ are the outputs of REB. $\hat{\mathbf{x}}$, $\hat{\mathbf{y}}$ represent the estimated values of $\mathbf{x}$ and $\mathbf{y}$, respectively; ${{\hat{\mathbf{x}}}_{t}}=[\hat{\mathbf{x}},\hat{\mathbf{y}}]$; $f(\cdot )$ is mapping function related to the inputs, ${{\phi }_{s}}$ is the parameters of the student network, ${{\theta }_{t}}$ is the parameters of the teacher network, ${{\phi }_{d}}$ is the parameters of the decorder network.

When the noise of the teacher-student network is evaluated, the robust teacher features are considered as the mixture of the noise and the features of the teacher network. Then, according to Eq. \ref{E6}, we maximize the mutual information between  the robust teacher features and the supervision information to learn the parameters that are most conducive to recovering input data  ${{\mathbf{x}}_{t}}$ from measured data $\mathbf{z}$ in the case of uncertainty. The mutual information between the representative features $\mathbf{z}$ and the supervision information ${{\mathbf{x}}_{t}}$ can be calculated as:

\begin{equation}
\label{E12}
{{I}_{{{\phi }_{s}}}}\mathbf{({{x}_{t}},z)}={H}(\mathbf{{{x}_{t}}})-{{{H}}_{{{\phi }_{s}}}}\mathbf{({{x}_{t}}\mid z)},
\end{equation}

\noindent where ${H}$ is the differential entropy. Thus, the optimization objective can be formulated as:

\begin{equation}
\begin{aligned}
\label{E13}
& {{\phi }_{s}}^{*}=\arg {{\max }_{{{\phi }_{s}}}}{{I}_{{{\phi }_{s}}}}\mathbf{({{x}_{t}},z)} \\ 
&=\arg {{\max }_{{{\phi }_{s}}}}-{{{H}}_{{{\phi }_{s}}}}\mathbf{({{x}_{t}}\mid z)}\\ 
&=\arg {{\max }_{{{\phi }_{s}}}}{\mathbf{\mathbb{E}}_{{{Q}_{{{\phi }_{s}}}}\mathbf{({{x}_{t}},z)}}}\left[ \log {{q}_{{{\phi }_{s}}}}({{\mathbf{x}}_{t}}\mid \mathbf{z}) \right]  
\end{aligned}.
\end{equation}

The virtual features $\mathbf{z}$ are used to replace the real features of the student network through REB. In the robust training of the teacher network in the training phase, the virtual robust teacher features are parameterized by the real teacher features $\mathbf{{{z}_{t}}}$ and an isotropic Gaussian noise according to Eq. \ref{E10}. The features from the student network are used to mimic the features from the teacher network:

\begin{equation}
\label{E14}
{{\phi }_{s}}^{*}=\arg {{\min }_{{{\phi }_{s}}}}{{\mathbb{E}}_{{{Q}_{{{\phi }_{s}},{{\phi }_{t}}}}\mathbf{({{X}_{t}},{{Z}_{t}},{{Z}_{s}})}}}\left[{\|{\mathbf{z}_{s}}-{\mathbf{z}_{t}}\|}_2^2\right].
\end{equation}

In this way, the performance of the student network is given back to the teacher network to train through uncertainty modeling on REB. The objective functions of the whole network are calculated through:

\begin{equation}
\begin{aligned}
\label{E15}
&{{\phi }_{s}}^{*}=\arg {{\min }_{{{\phi }_{s}}}}{{\mathbb{E}}_{{{Q}_{{{\phi }_{s}},{{\phi }_{t}}}}\mathbf{({{X}_{t}},{{Z}_{t}},{{Z}_{s}})}}}\left[ \|{{\mathbf{z}}_{s}}-{{\mathbf{z}}_{t}}\|_{2}^{2} \right] \\ 
& {{\phi }_{t}}^{*}=\arg {{\min }_{{{\phi }_{t}}}} {{\mathbb{E}}_{{{Q}_{{{\phi }_{s}},{{\phi }_{t}}}}\mathbf{({{X}_{t}},{{Z}_{t}},{{Z}_{s}})}}}\left[ \log {{q}_{{{\phi }_{t}}}}({{\mathbf{x}}_{t}}\mid \mathbf{z}) \right] \\ 
& s.t.\qquad{z}\sim\mathcal{N}\left( {{\mathbf{z}}_{t}},{{\sigma }^{2}}\mathbf{I} \right)
\end{aligned}.
\end{equation}

We adopt the parameterized variational approximation of the posterior probability ${{q}_{{{\phi }_{t}}}}({{\mathbf{x}}_{t}}\mid \mathbf{z})$ as ${{p}_{{{\theta }_{d}}}}({{\mathbf{x}}_{t}}\mid \mathbf{z})$, and thus the loss of the teacher network can be calculated as:

\begin{equation}
\begin{aligned}
\label{E16}
& {{\mathcal{L}}_{t}}={{\mathbb{E}}_{{{Q}_{{{\phi }_{s}},{{\phi }_{t}}}}\mathbf{({{X}_{t}},{{Z}_{t}},{{Z}_{s}})}}}\left[ \left\| {{\mathbf{x}}_{t}}-{{f}_{d}}(\mathbf{z};{{\theta }_{d}}) \right\|_{2}^{2} \right] \\ 
&s.t.\qquad{z}\sim\mathcal{N}\left( {{\mathbf{z}}_{t}},{{\sigma }^{2}}\mathbf{I} \right)
\end{aligned},
\end{equation}

\noindent The loss of the student network can be calculated through:

\begin{equation}
\begin{aligned}
\label{E17}
&{{\mathcal{L}}_{s}}={{\mathbb{E}}_{{{Q}_{{{\phi }_{s}},{{\phi }_{t}}}}\mathbf{({{X}_{t}},{{Z}_{t}},{{Z}_{s}})}}}\left[ \|{{\mathbf{z}}_{s}}-{{\mathbf{z}}_{t}}{{\|}^{2}} \right] \\
&s.t.\qquad{z}\sim\mathcal{N}\left( {{\mathbf{z}}_{t}},{{\sigma }^{2}}\mathbf{I} \right)
\end{aligned}.
\end{equation}

As $\mathcal{L}_t$ and $\mathcal{L}_s$ cannot be optimized analytically, asynchronous iteration is adopted in the training process of optimizing the objective function of the two networks. We asynchronously iterate ${{\phi }_{t}}$ and ${{\phi }_{s}}$ according to Eqs. \ref{E16} and \ref{E17} when ${{\sigma }^{2}}$ is evaluated. ${{\sigma }^{2}}$ is the intermediate variable in the iterative training process, which is updated according to Eq. \ref{E12} at the end of each iteration. When ${{\mathcal{L}}_{t}}$ and ${{\mathcal{L}}_{s}}$ reach the set value or do not change significantly, it can be considered that the teacher-student network has been trained to extract the maximum manual information, that is, the quality-relevant features most conducive to the recovery of the process variables have been extracted to the maximum extent.

The overall algorithms of TSUAE are described in algorithm 1 and algorithm 2. 

\vskip 0in
\begin{algorithm}[H]
	\caption{The training phase for teacher-student uncertainty autoencoder}
	\begin{algorithmic}[1] 
		\Require $\theta_{s}$, $\theta_{t}$, $f_{t}(\cdot)$, $f_{d}(\cdot)$, $f_{s}(\cdot)$, $\sigma^2$, $opt_s$, $opt_t$, stop criterion
		\Ensure  $f_{t}(\cdot)$, $f_{d}(\cdot)$, $f_{s}(\cdot)$
		\State \textbf{while} stop criterion is not met \textbf{do}
		\State
		\hspace*{0.8cm} Sample: $\mathbf{x}_{t}, \mathbf{x}_{s} \sim X_{t}, X_{s}$ 
		
		\State 	\hspace*{0.8cm} Calculate $\mathbf{z}_{t}$,$\mathbf{z}_{s}$ according to: $\mathbf{z}_{t}=f_{t}\left(\mathbf{x}_{t}\right)$,\hspace{0.5cm}$\mathbf{z}_{s}=f_{s}\left(\mathbf{x}_{s}\right)$
		\State
		\hspace*{0.8cm} Sample:	 $\mathbf{d}_f \sim \mathcal{N}\left( 0,{{\sigma }^{2}}I \right)$
		\State \hspace*{0.8cm} Calculate	$\mathbf{z}$ according to: $\mathbf{z}=\operatorname{REB}\left(\mathbf{z}_{t}, \mathbf{z}_{s}, \text{\textit{phase}}\right)=\mathbf{z}_{t}+\mathbf{d}_f $
		\State \hspace*{0.8cm} Calculate $\hat{\mathbf{x}}_{t}$ according to: $\hat{\mathbf{x}}_{t}=f_{d}(\mathbf{z})$
		\State \hspace*{0.8cm} Calculate $\mathcal{L}_{t}$ according to: $\mathcal{L}_{t}=\mathbb{E}\left[\left\|\mathbf{x}_{t}-f_{d}\left(\mathbf{z} ; \theta_{d}\right)\right\|_{2}^{2}\right]$
		\State \hspace*{0.8cm} Calculate $\mathcal{L}_{s}$  according to: $\mathcal{L}_{s}=\mathbb{E}\left[ {\|{{\bf{z}}_s} - {{\bf{z}}_t}\|_2^2} \right]$
		\State \hspace*{0.8cm} Update $\sigma^2$ according to: ${\sigma ^2} = \mathbb{E}\left[ {\|{{\bf{z}}_t} - {{\bf{z}}_s}\|_2^2} \right]$
		\State   \hspace*{0.8cm} Execute one optimizing step for $\phi_{s}$: $\phi_{s}=opt_s(\phi_{s},\nabla_{\phi_s} \mathcal{L}_s)$
		\State   \hspace*{0.8cm} Execute one optimizing step for $\phi_{t},\theta_{d}$:\\
		\hspace*{2.5cm}$\left\{\phi_{\mathrm{t}},\theta_{d}\right\}={opts}\left(\left\{\phi_{\mathrm{t}}, \theta_{d}\right\}, \nabla_{\left\{\phi_{t}, \theta_{d}\right\}} \mathcal{L}_{t}\right)$
		\State \textbf{End while}
		\State \textbf{Return} $f_{t}(\cdot)$,$f_{d}(\cdot)$,$f_{s}(\cdot)$

	\end{algorithmic}
\end{algorithm}


\vskip 0in
\begin{algorithm}[H]
	\caption{The testing phase for teacher-student uncertainty autoencoder}
	\begin{algorithmic}[1] 
		\Require $\mathbf{x}_{s}$, $f_{t}(\cdot)$, $f_{d}(\cdot)$, $f_{s}(\cdot)$
		\Ensure  $\hat{\mathbf{x}}_{t}$
		
		\State 	Calculate $\mathbf{z}_{s}$ according to: $\mathbf{z}_{s}=f_{s}\left(\mathbf{x}_{s}\right)$
		
		\State Calculate	$\mathbf{z}$ according to: $\mathbf{z}=\operatorname{REB}\left(\mathbf{z}_{t}, \mathbf{z}_{s}, \text{phase}\right)=\mathbf{z}_{s}$
		\State Calculate $\hat{x}_{t}$ according to: $\hat{\mathbf{x}}_{t}=f_{d}(\mathbf{z})$
		
		\State \textbf{Return} $\hat{\mathbf{x}}_{t}$

	\end{algorithmic}
\end{algorithm}

The teacher-student network is trained in the training phase through algorithm 1. Specifically, the parameters of the network are initialized and the initial hyperparameter of the standard deviation is set first. In each iteration, we sample the noise from the current noise condition and calculate the robust teacher features according to Eq. \ref{E9}. Then the teacher network is updated according to the loss function calculated from the robust teacher features according to Eq. \ref{E16}, and the student network is supervised according to the loss function calculated from the representative features of REB according to Eq. \ref{E17}. Then the noise condition is updated according to Eq. \ref{E13}. Thus, the parameters of the teacher-student network can be updated. Finally, the iteration process repeated until the stop condition is satisfied. In the testing phase, TSUAE is used to get the representative student features from the student network when TSUAE has been trained through algorithm 2.

At last, the complexity of the training algorithm is calculated. Assuming that the number of iterations of the teacher-student network in the training process is $N$ and the algorithm complexity of each iteration is $O(T)$, the algorithm complexity is related to the network structure and can be simplified as $O(T)=O(n_{in}n_{z}+n_{z}n_{h}+n_{h}n_{out})$ for the proposed teacher-student structure in the paper, where $n_{in}$ is the number of inputs of the structure; $n_{z}$ is the number of features of the structure; $n_{h}$ is the number of neurons in the hidden layer of the structure; $n_{out}$ is the number of outputs of the structure. Thus, the algorithm complexity of the training phase is $O(NT)$ while the algorithm complexity of the testing phase is $O(T)$. Therefore, our proposed method is effective in computational complexity in the training phase and the testing phase.

\subsection{Fault detection method based on teacher-student uncertainty autoencoder}

Considering the high cost and time delay of measuring the quality-relevant variables, it becomes relatively difficult, even in some cases, impractical to monitoring the fault occurs in quality-relevant subspace. The literature \cite{Yan2019} indicated that the fault detection in quality-relevant subspace can be realized through the prediction of quality-relevant variables. Accordingly, TSUAE is used to extract the quality-relevant representative features by mimicking the teacher features with process-relevant variables and quality-relevant variables. Thereby TSUAE can be used to predict the  process-relevant variables and the quality-relevant variables:

\begin{equation}
\label{E18}
\mathbf{{{\hat{x}}_{t}}}=[\mathbf{\hat{x},\hat{y}}],
\end{equation}

\noindent where $\mathbf{\hat{x}}$ is the estimated value of the process-relevant variables, and $\mathbf{\hat{y}}$ is the estimated value of the quality-relevant variables. Therefore, two subspaces representing the process-relevant error and the quality-relevant error are designed. On this basis, two error statistics are constructed in these two subspaces:

\begin{equation}
\label{E19}
{{D}_{\mathbf{x},i}}=\|{{\mathbf{\hat{x}}}_{i}}-{{\mathbf{x}}_{i}}\|_{2}^{2},
\end{equation}

\begin{equation}
\label{E20}
{{D}_{\mathbf{y},i}}=\|{{\mathbf{\hat{y}}}_{i}}-{{\mathbf{y}}_{i}}\|_{2}^{2},
\end{equation}

\noindent where $i$ is the index of samples of the data.

The threshold is determined by ${{D}_\mathbf{x}}$ and ${{D}_\mathbf{y}}$ calculated from the training samples, and ${{J}_{\mathbf{x},th}}$ and ${{J}_{\mathbf{y},th}}$ are determined by KDE. For the ${{D}_{\mathbf{x},new}}$ and ${{D}_{\mathbf{y},new}}$ are calculated from the new sample ${\mathbf{x}_{\mathbf{t},new}}$, if ${{D}_{\mathbf{x},new}}$ exceeds ${{J}_{\mathbf{x},th}}$, the fault can be considered to be detected. Furthermore, if ${{D}_{\mathbf{y},new}}$ exceeds ${{J}_{\mathbf{y},th}}$, a quality-relevant fault can be considered to have occurred.

In the offline robust training of the teacher network, process-relevant variables $\mathbf{x}$ and quality-relevant variables $\mathbf{y}$ are used to calculate teacher features $\mathbf{{{z}_{t}}}$ and student features $\mathbf{{{z}_{s}}}$. The outputs $\mathbf{z}$ calculated from the REB are considered as the robust features of the teacher network. The robust features of the teacher network are the mixture of the teacher features and an isotropic Gaussian noise representing the inconsistency of the two networks. In online process monitoring when the quality variables are unavailable, the teacher features $\mathbf{{{z}_{t}}}$ are unknowable. REB is used to get real student features in the testing phase.  Thus, the outputs of the testing phase can be used to evaluate the process-relevant variables and quality-relevant variables. After that, the error and corresponding threshold are calculated in the process-relevant and quality-relevant subspace in offline modeling when quality-relevant variables are available.

In the online application, the pre-trained student model is used to calculate student features $\mathbf{{{z}_{s,new}}}$ from online process-relevant variables $\mathbf{{{x}_{new}}}$ first. Next, REB is used to get the corresponding outputs $\mathbf{{{z}_{new}}}$, which are also the student features when the REB in the testing phase. After that, the corresponding outputs are used to calculate the evaluated process-relevant variables $\mathbf{{{\hat{x}}_{new}}}$ and the evaluated quality-relevant variables $\mathbf{{{\hat{y}}_{new}}}$, which are then used to calculate the corresponding threshold  ${{D}_{\mathbf{x},new}}$ and ${{D}_{\mathbf{y},new}}$. Finally, the quality-relevant subspace monitoring is realized through the monitoring decision in the process-relevant subspace and generates a subsequent evaluation of the analysis for the impact of fault on quality-relevant variables when the quality-relevant variables are available.

\begin{figure}[t]
	\vskip 0in
	\begin{center}
		\centerline{\includegraphics[width=\columnwidth]{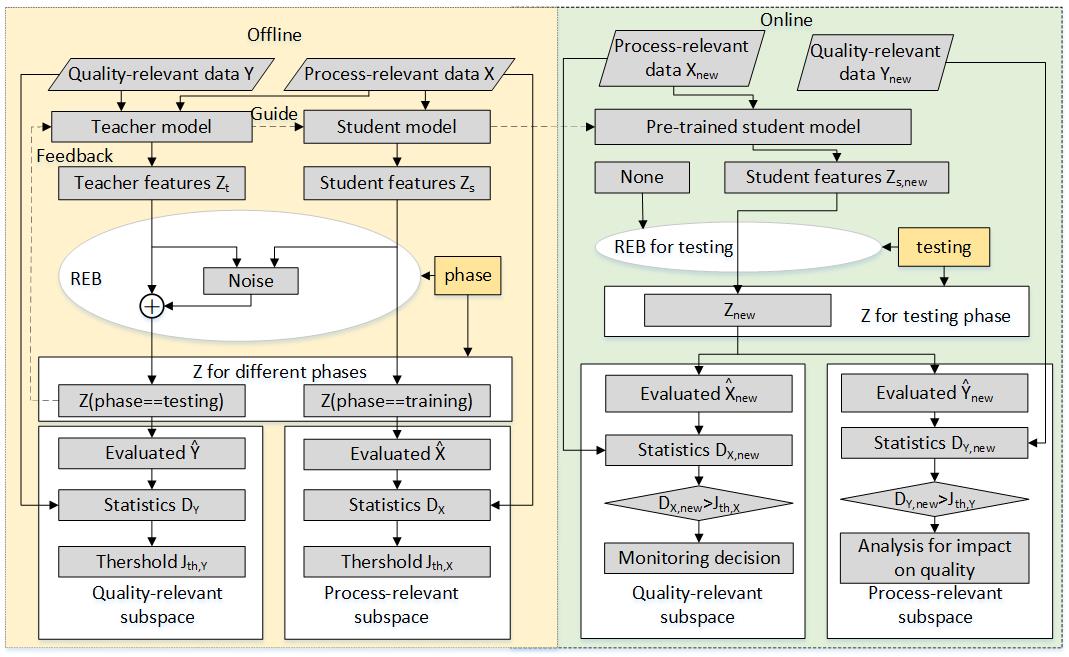}}
		\caption{The flowchart of the TSUAE-based quality-relevant fault detection method}
		\label{fig4}
	\end{center}
	\vskip -0.4in
\end{figure}

The proposed process monitoring method based on TSUAE mainly contains two parts: offline modeling and online monitoring, which is shown in Fig. \ref{fig4}.

\section{Experiments}
To evaluate the performance of the proposed method, the proposed method is verified in the numerical example and the BSM1 simulation and evaluated through  false alarm rate (FAR) and the fault detection rate (FDR), which are defined as:
\begin{equation}
\label{E21}
\mathrm{FAR}=\frac{\mathrm{N}_{\mathrm{fa}}}{\mathrm{N}_{\mathrm{n}}} \times 100 \% ,
\end{equation}
\begin{equation}
\label{E22}
\mathrm{FDR}=\frac{\mathrm{N}_{\mathrm{fd}}}{\mathrm{N}_{\mathrm{f}}} \times 100 \% ,
\end{equation}

\noindent where  $\mathrm{N}_{\mathrm{n}}$ and $\mathrm{N}_{\mathrm{f}}$  denote the number of samples under normal condition and abnormal condition, respectively. $\mathrm{N}_{\mathrm{fa}}$ and $\mathrm{N}_{\mathrm{fd}}$ represent the number of samples that issue false alarm and detect faults, respectively.

Our proposed method is compared with fault detection methods based on PCA \cite{Fan2014}, PLS \cite{Wilson2000}, ridge regression(RR) \cite{Zhang2021}, SAE \cite{Zheng2020} and TSSAE \cite{Yan2019} in the process-relevant subspace or in the quality-relevant subspace. PCA and PLS are widely used when inspecting process failures, while RR adds a regularization term, which has a better effect on overfitting and ill-conditioned data. SAE and TSSAE are fault detection by the neural network, and TSSAE uses the teacher-student network. Compared with these methods, the performance of our method can be comprehensively evaluated.

\subsection{numerical example}

To illustrate the performance of our proposed monitoring method based on TSUAE, a numerical example is designed. The variables in this example are composed of latent variables mapped from latent variables with noise. The quality-relevant variables are related to some of these latent variables and used to validate the ability to extract quality-relevant features. In this example, six input variables are generated from the same standard Gaussian distribution and unit variance. The output variable is set to a nonlinear function of ${{z}_{1}}, {{z}_{2}}, {{z}_{3}}, {{z}_{4}}, {{z}_{5}}$, as is shown below:

\begin{equation}
\begin{aligned}
\label{E23}
&{{z}_{i}}\sim N(0,1)\text{   }i=1,2,3,4,5,6\\ 
&\mathbf{x}=\mathbf{W}\mathbf{z}+\mathbf{e_{i}}\text{       ,} \mathbf{e_{i}}\sim \mathcal{N}(0,0.1\textbf{I})\text{ ,}{{w}_{i,j}}\sim \mathcal{N}(0,1) \\ 
& \mathbf{y}={{({{z}_{1}}+{{z}_{2}})}^{2}}+\exp (({{z}_{3}}-{{z}_{4}})/2)+\sin ({{z}_{5}})
\end{aligned},
\end{equation}

\noindent where $\mathbf{W}:{{\mathbf{R}}^{6}}\to {{\mathbf{R}}^{20}}$ is a linear mapping from $\mathbf{z}$ to $\mathbf{x}$, whose elements are independent and follow the standard normal distribution, $\mathbf{x}=[{{x}_{1}},{{x}_{2}},\cdots {{x}_{20}}]$ are the input variables, $\mathbf{z}=[{{z}_{1}},{{z}_{2}},\cdots ,{{z}_{6}}]$ are the latent variables, $\mathbf{y}=[y]$ is
the quality-relevant variable. The activation functions of the recovery information are nonlinear. Both the training samples and the two fault samples are sampled 1000 times under this condition. The representative features of two networks are extracted by linear functions where the recovery of the combined data is mapped with a two-layer neural network with Tanh as a non-linear activation function. The faults are constructed as follows:

\textbf{Fault 1}: A step fault increased by 1.5 from the 201st sample of the 10th input variable;

\textbf{Fault 2}: A step fault increased by 0.7 from the 201st sample of the quality-relevant indicator.

\begin{figure}[]
	\vskip 0in
	\begin{center}
		\centerline{\includegraphics[width=\columnwidth]{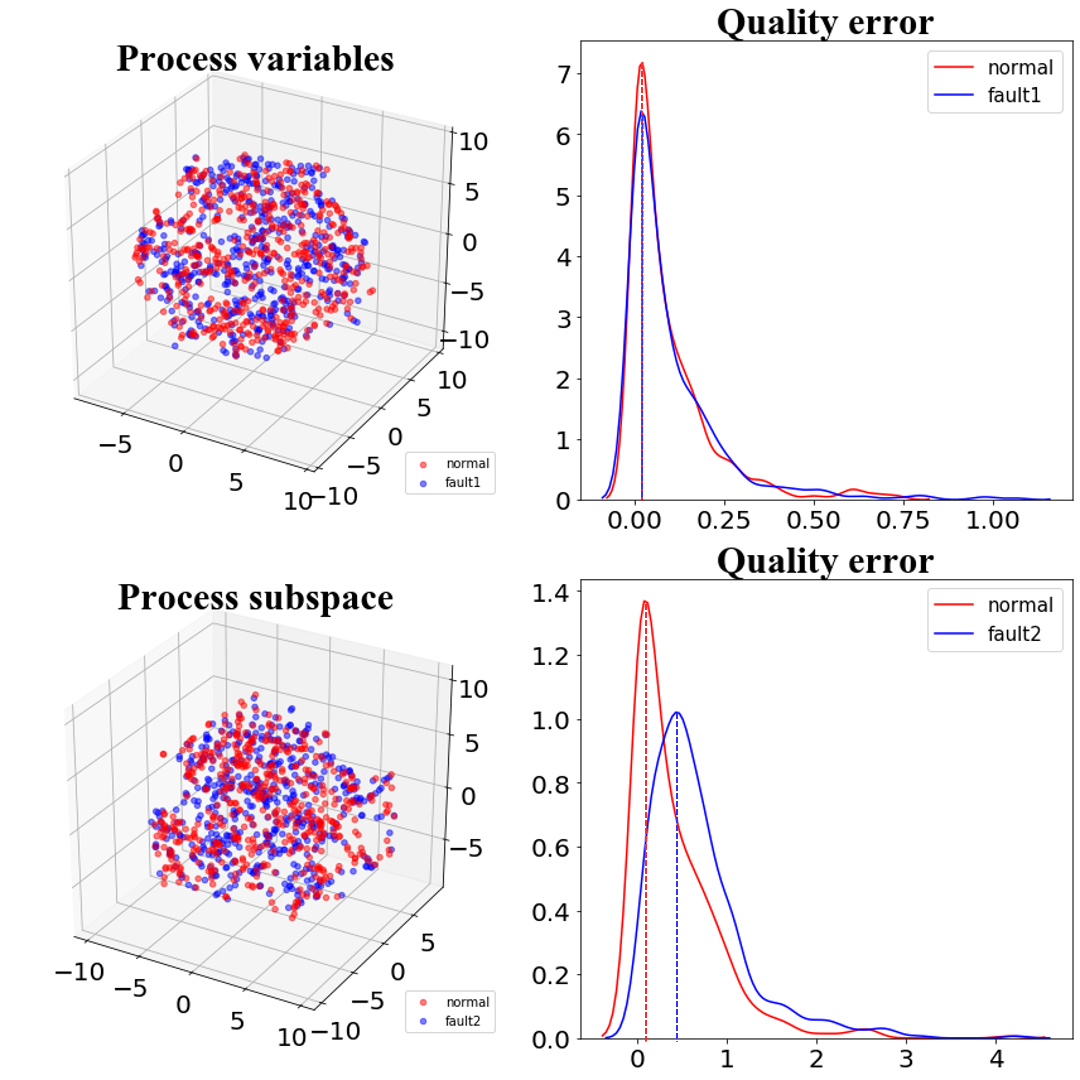}}
		\caption{Visualization of faults in numerical example}
		\label{fig5}
	\end{center}
	\vskip -0.4in
\end{figure}

To analyze the features of the process-relevant subspace and the quality-relevant subspace, the process-relevant subspace and the quality-relevant subspace are visualized in Fig. \ref{fig5}, respectively. In the process-relevant subspace which contains a lot of process variables, t-distributed stochastic neighbor embedding (t-SNE) \cite{Laurens2008}, a nonlinear dimension reduction algorithm, is used to reduce the dimension of data to the three-dimensional and then visualized the three-dimensional scatterplot in the left of Fig. \ref{fig5}. In the quality-relevant subspace, the relationship between the process variables and the quality indicators are established through PLS according to the normal data. After that, the KDEs of the residual between the quality indicators and the evaluation calculated from the pre-trained PLS model are recorded in the right of  Fig. \ref{fig5}.

In the fault 1 and the fault 2, the differences between normal data and fault data cannot be recognized from scatterplot based on t-SNE. However, the process variables of fault 1 have changed according to the process abnormality. This is because t-SNE reduces the change of the process abnormality through dimensional reduction when maintaining the similarity between the data. In the quality subspace, the fault 1 is considered to be a quality-irrelevant fault which reflects the similarity of the KDEs between the normal data and the abnormal data. As the fault 2 is a quality-relevant fault, the KDEs between the normal data and the abnormal data are obviously different. However, the KDEs of the normal data and the abnormal data are overlapped to a certain extent, and thus the detection method based on the residual subspace of PLS contains a large fault missing rate.

\begin{figure}[t]
	\vskip 0in
	\begin{center}
		\centerline{\includegraphics[width=\columnwidth]{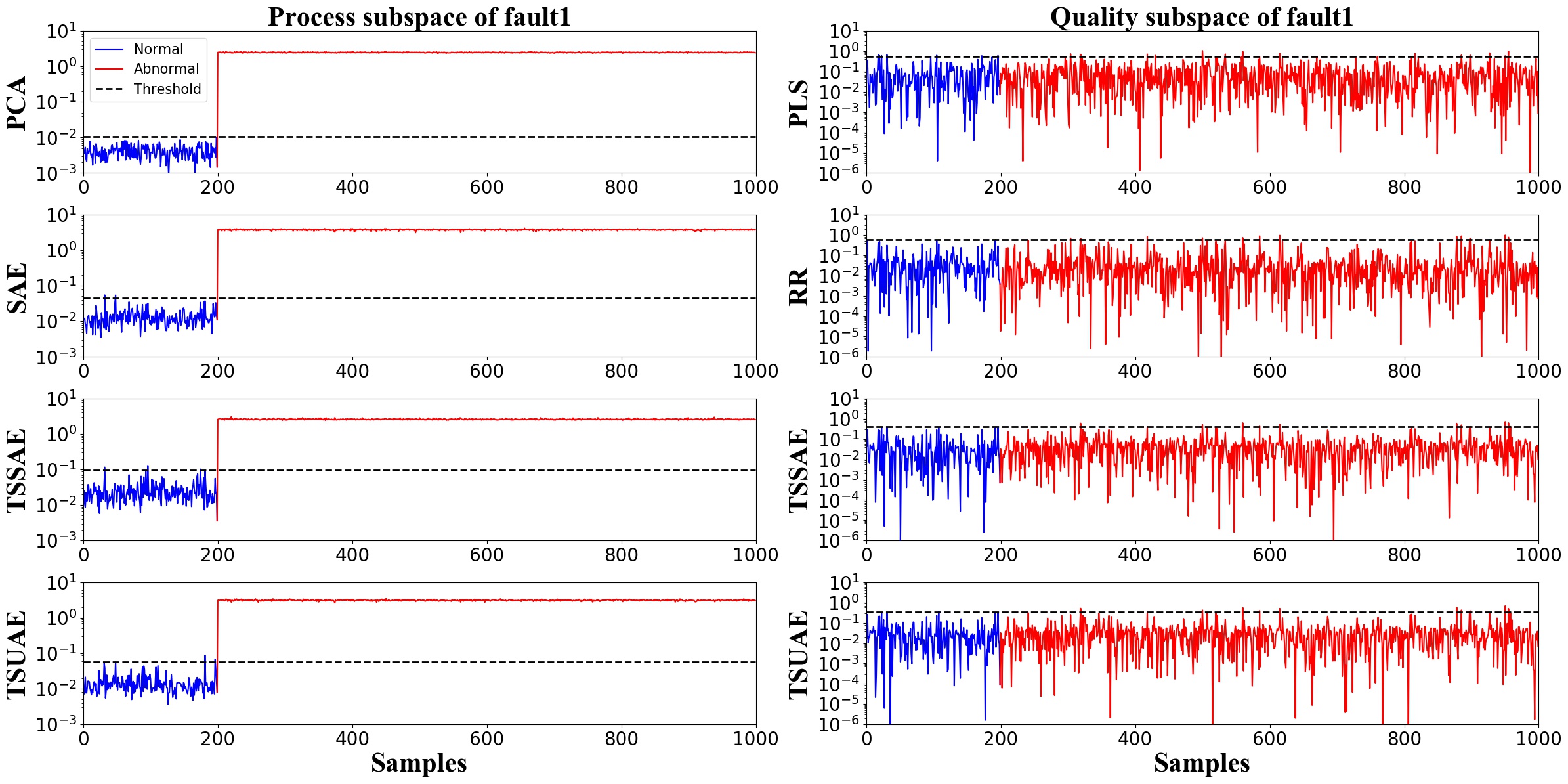}}
		\caption{Fault detection performance of fault 1 in the numerical example}
		\label{fig6}
	\end{center}
	\vskip -0.4in
\end{figure}

\begin{figure}[t]
	\vskip 0in
	\begin{center}
		\centerline{\includegraphics[width=\columnwidth]{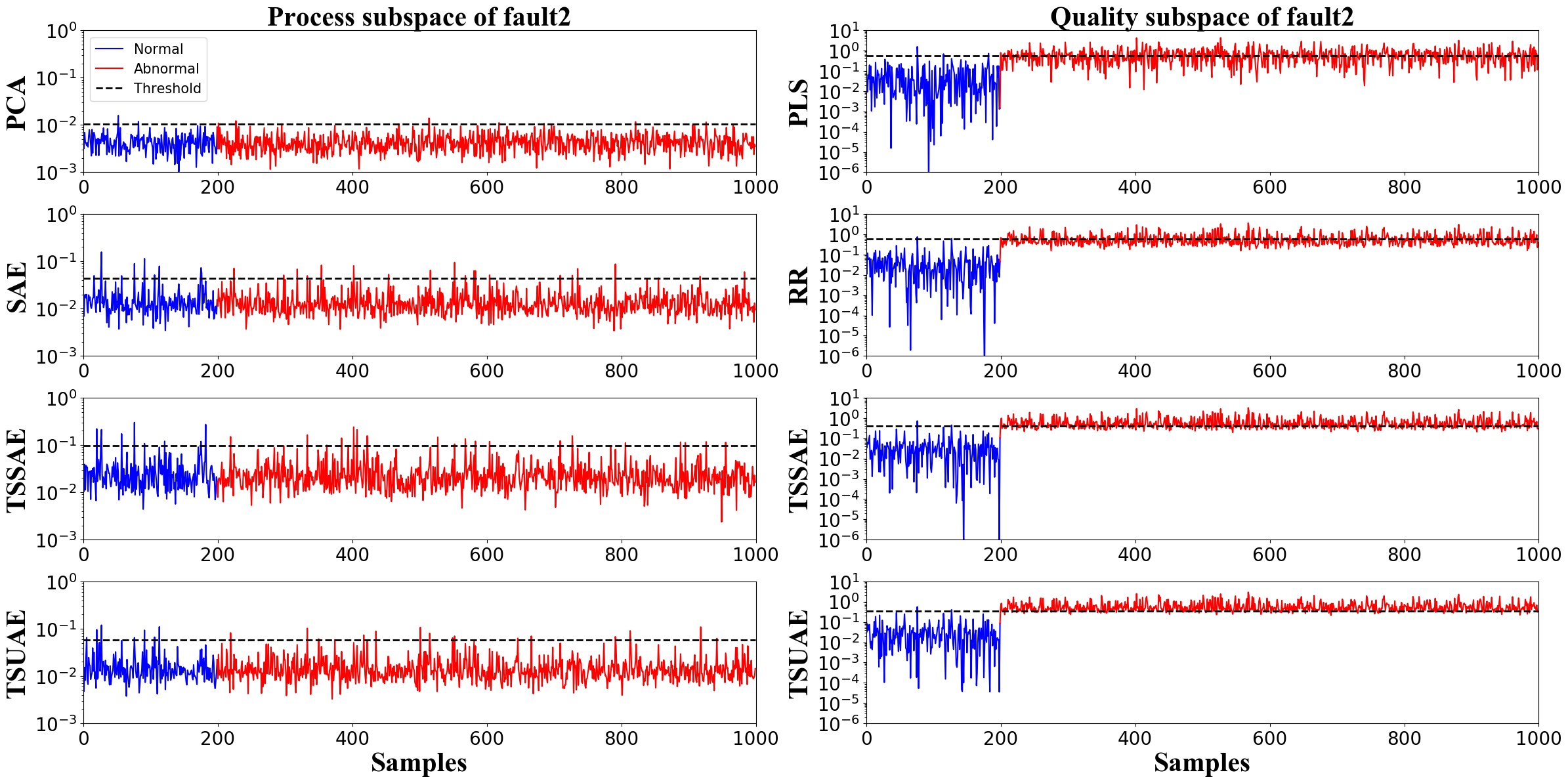}}
		\caption{Fault detection performance of fault 2 in the numerical example}
		\label{fig7}
	\end{center}
	\vskip -0.4in
\end{figure}

The result of fault 1 is shown in Fig. \ref{fig6}, and the result of fault 2 is shown in Fig. \ref{fig7}. The FARs and FDRs of detection results are summarized in Table \ref{table1}.  

\begin{table}[!htb]
	\caption{FARs/FDRs for numerical example in process-relevant subspace and quality-relevant subspace}
	\label{table1}
	\begin{center}
		\begin{tabular}{@{}ccccc@{}}
			\hline
			&&&& \\[-10pt] 
		& \multicolumn{2}{c}{1}      & \multicolumn{2}{c}{2} \\ \cmidrule(l){2-5} 
			& Dx         & Dy            & Dx    & Dy            \\ \midrule
			\textbf{TSUAE}                      & 0.01/\textbf{1.00}                    & 0.01/0.01                   & 0.04/0.02                    &\textbf{0.01}/ \textbf{0.78} \\
			\textbf{TSSAE}                       & 0.02/\textbf{1.00}                    & \textbf{0.00}/0.01                    & 0.05/\textbf{0.03}                    &\textbf{0.01}/ 0.62 \\
			\textbf{SAE}                        & 0.01/\textbf{1.00}             & /                       & 0.05/\textbf{0.03}                   & /    \\
					\textbf{PCA}                        & \textbf{0.00}/\textbf{1.00}                    & /                       & \textbf{0.01}/0.01                    & /    \\
			\textbf{PLS}                        & /                    & 0.04/0.01                       & /                    & 0.02/0.47    \\
			\textbf{RR}                       & /                       &0.01/ \textbf{0.02 }                   & /                     & \textbf{0.01}/0.40 \\

			\hline       
		\end{tabular}
	\end{center}
\end{table}

The thresholds are determined by KDE with the 99\% confidence interval, so the FARs of all methods are small. In this case, we focus on analyzing the FDR of each method. The fault 1 occurs to the process-relevant variables, which has little impact on the prediction of the quality-relevant subspace. Therefore, the fault detection methods based on the process-relevant subspace can detect the fault, while fault detection methods based on quality-relevant subspace cannot detect the fault. It is worth mentioning that the fault detection method based on the statistics of the process variables can detect the abnormality that can not be distinguished by the visualization method, representing the effectiveness of fault detection methods based on the statistics.
The fault 2 is a quality-relevant fault. As the process-relevant variables do not contain abnormality, the fault detection methods based on process-relevant subspace cannot detect the fault, but the quality-relevant fault detection methods such as TSSAE and TSUAE can detect the fault. PLS, RR and TSSAE cannot detect faults in the most time, while our proposed TSUAE can effectively detect faults in the most time. The numerical example proved the model of TSUAE describing the relationship between quality information and input variables was more accurate than others. 

A lower threshold means higher fault sensitivity. The thresholds of different methods are shown in Table \ref{table2}, which are determined by KDE with the 99\% confidence interval. The subspace of the TSUAE outputs has a lower threshold value, indicating that it better captures the relationship between input and quality-relevant information. Moreover, the method focuses more on the ability to recover the quality-relevant information because it extracts the quality-relevant representative features for process-relevant variables and combination variables of quality-relevant and process-relevant. Although the threshold of the process-relevant subspace is larger than others, it can still accurately detect the faults. Since the process subspace of this simulation is a linear model from the hidden variables, PCA based on the linear model has the lowest threshold of the process-relevant subspace.

\begin{table}[!htb]
	\caption{The thresholds of different methods in the numerical example}
	\label{table2}
	\begin{center}
		\begin{tabular}{@{}ccccccc@{}}
			\hline
			&&&& \\[-10pt] 
			 & TSUAE & TSUAE & SAE & PCA & PLS & RR \\ 
			\hline
			Dx & 0.078 & 0.095 &0.044 & \textbf{0.010} & / & / \\ 
			Dy & \textbf{0.200} & 0.398 & / & / & 0.570 & 0.548 \\
			\hline       
		\end{tabular}
	\end{center}
\end{table}

\begin{figure}[t]
	\vskip 0in
	\begin{center}
		\centerline{\includegraphics[width=\columnwidth]{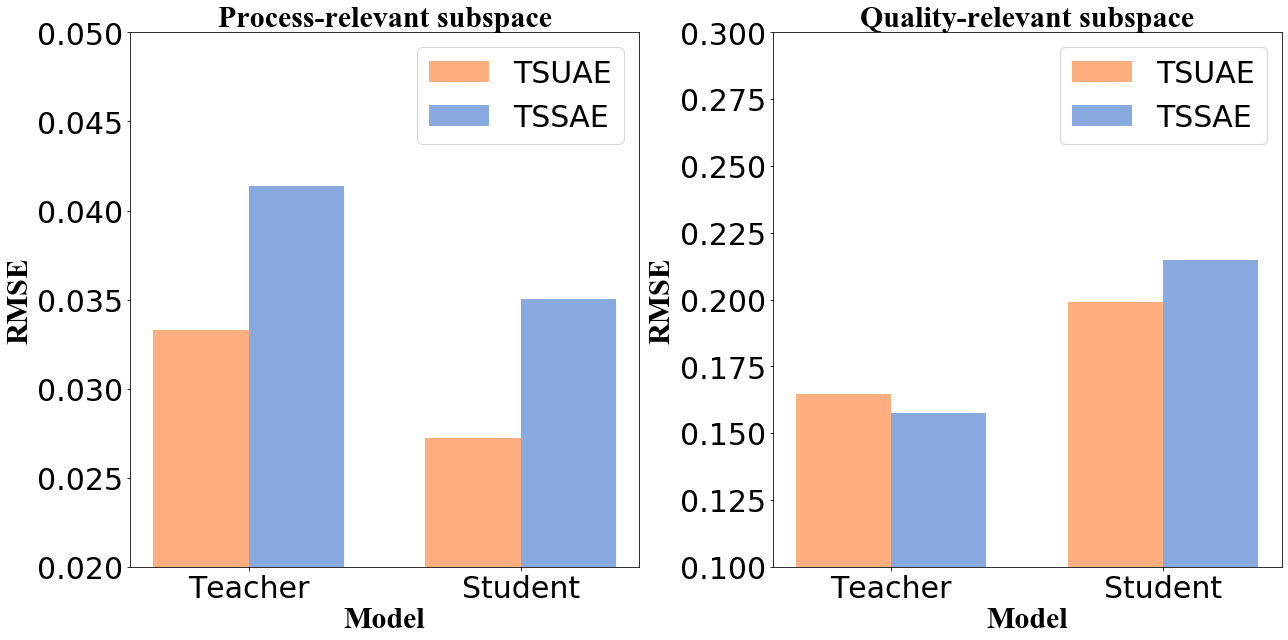}}
		\caption{The error of replacing the teacher network with the student network in different subspaces}
		\label{fig8}
	\end{center}
	\vskip -0.4in
\end{figure}

To verify the performance of the TSUAE, we record the root-mean-square errors of the teacher network and the student network in the process-relevant subspace and quality-relevant subspace during the training process, respectively, as is shown in Fig. \ref{fig8}. In the process-relevant subspace, because of the large number of variables, the noise feedback in TSUAE improves the robustness of the teacher network, so our proposed TSUAE contains a lower error, which implies that the model of TSUAE has better features extraction capabilities. In the quality-relevant subspace with only one quality-relevant variable, the performance of the teacher network in TSUAE is worse than in TSSAE after introducing perturbation. However, when the features of the student network are used to replace the features of the teacher network, the prediction performance of TSUAE is better than TSSAE. On the whole, replacing the features of the teacher network with that of the student network helps the recovery of the process-relevant variables. At the same time, it will increase the prediction error of the quality-relevant variables. The reason is that the inputs of the student network are process-relevant variables, which are conducive to process recovery, but lack of representability of quality-relevant variables. Thus it can be inferred that the optimal teacher model is conducive to the prediction of quality variables, but the features differences will reduce the predicting performance dramatically. Our proposed TSUAE can reduce performance degradation on the quality-relevant variables.

\subsection{Analysis of the proposed method}

\begin{figure}[t]
	\vskip 0in
	\begin{center}
		\centerline{\includegraphics[width=\columnwidth]{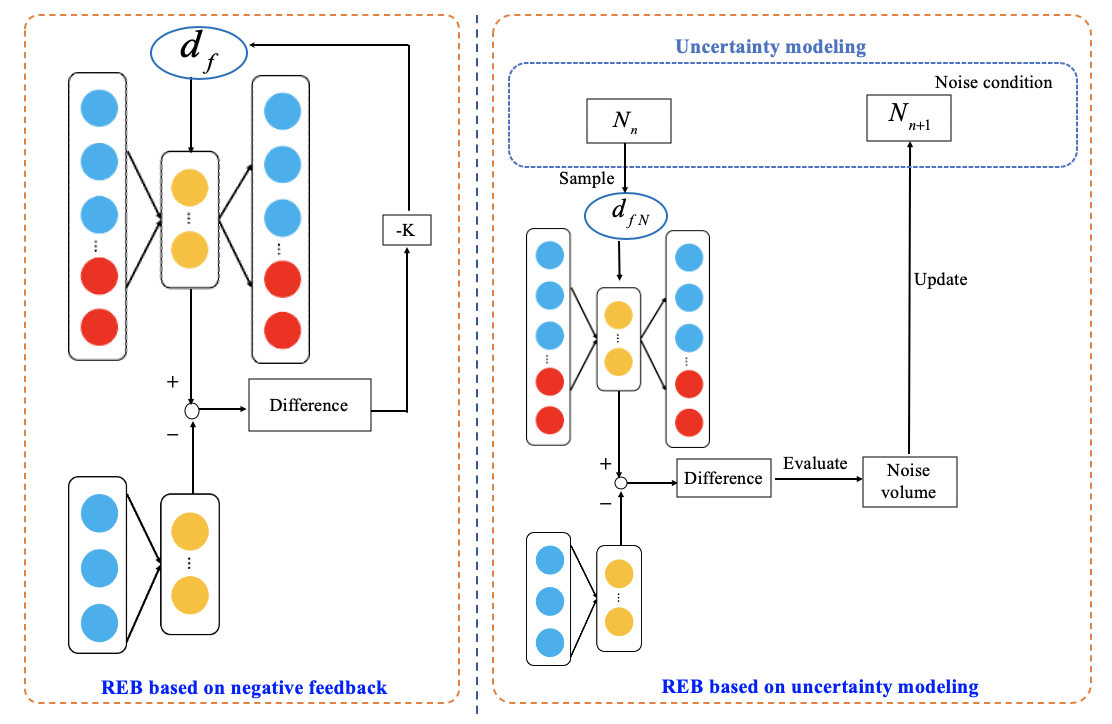}}
		\caption{The training processes of REB based on negative feedback and REB based on uncertainty modeling}
		\label{fig9}
	\end{center}
	\vskip -0.4in
\end{figure}

In order to analyze the role of REB and uncertainty modeling independently, REB is added in TSSAE with a conventional strategy in which the feedback is the features differences scaled by pre-set feedback rate $-k$, where $k$ is non-negative. Thus, the training processes of the REB based on negative feedback and REB based on uncertainty modeling are shown in Fig. \ref{fig9}. $N_{n}$ represents the noise based on the differences from the $Nth$ iteration in uncertainty modeling.

Therefore, the feedback of features differences can be calculated through:

\begin{equation}
\label{E24}
	\mathbf{d}_{f}=-k\left(\mathbf{z_{t}}-\mathbf{z_{s}}\right) .
\end{equation}

Therefore, the training process of REB based on negative feedback can be expressed as:
\begin{equation}
\label{E25}
\operatorname{REB}\left(\mathbf{z}_{t}, \mathbf{z}_{s},^{\prime} \text {training}^{\prime}\right)=\mathbf{z}_{t}-k\left(\mathbf{z}_{t}-\mathbf{z}_{s}\right)=(1-k) \mathbf{z}_{t}+k \mathbf{z}_{s} .
\end{equation}

\begin{figure}[]
	\vskip 0in
	\begin{center}
		\centerline{\includegraphics[width=\columnwidth]{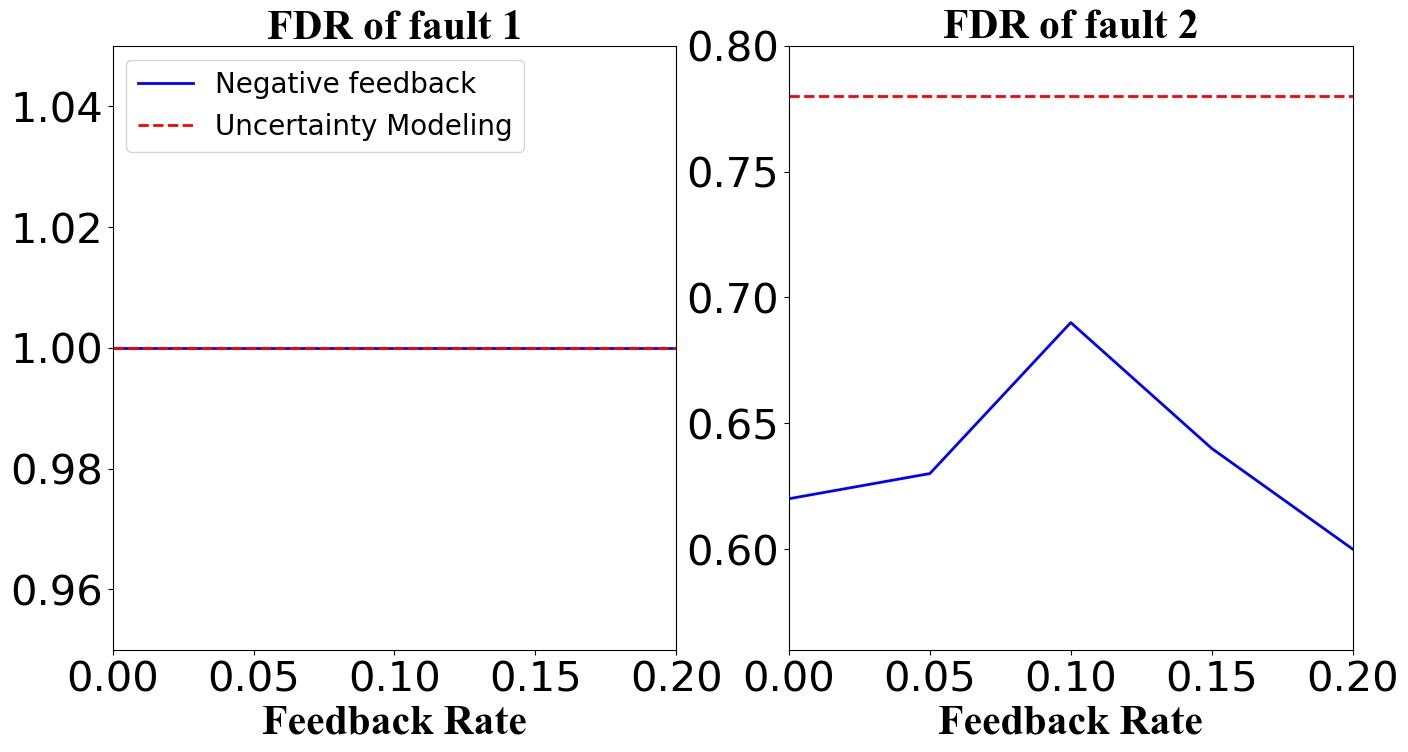}}
		\caption{The FDRs of different models under different feedback rates}
		\label{fig10}
	\end{center}
	\vskip -0.4in
\end{figure}

\begin{figure}[]
	\vskip 0in
	\begin{center}
		\centerline{\includegraphics[width=\columnwidth]{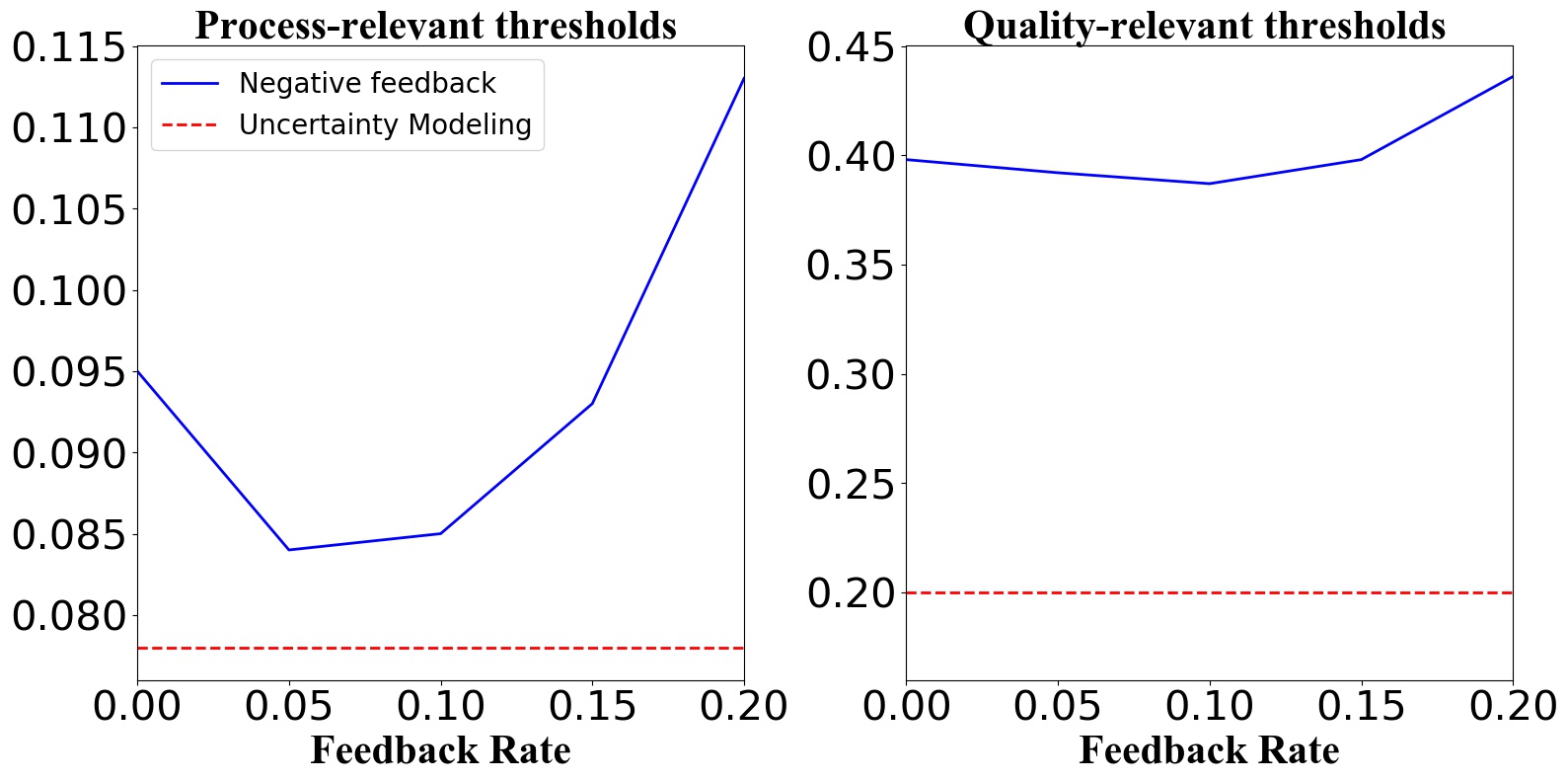}}
		\caption{The thresholds of different models under different feedback rates}
		\label{fig11}
	\end{center}
	\vskip -0.4in
\end{figure}

To verify the effectiveness of the REB and the proposed uncertainty modeling on REB,  the REB with a negative feedback rate is added to the TSSAE training process. In TSSAE,  the feedback is zero as the teacher network is pre-trained. Since the fault 1 is a process-relevant fault and the fault 2 is a quality-relevant fault, we mainly discuss the detection results of the corresponding subspace. Table \ref{table3} shows the FDRs and thresholds at different feedback rates. Figs. \ref{fig10}, \ref{fig11} show the FDRs and thresholds of model based on negative feedback and uncertainty under different feedback rates.

\begin{table}[!htb]
	\caption{The fault dection results with differernt negative rate in TSSAE}
	\label{table3}
	\begin{center}
	
	\begin{tabular}{@{}ccccc@{}}
		\toprule
		Negative rate & \multicolumn{2}{c}{Fault 1} & \multicolumn{2}{c}{Fault 2}  \\ \cmidrule(l){2-5}
		& FDR       & Threshold       & FDR        & Threshold      \\ \midrule
		0             & 1         & 0.0950          & 0.620      & 0.398          \\
		0.05          & 1         & \textbf{0.0844 }         & 0.630      & 0.392          \\
		0.1           & 1         & 0.0848          & \textbf{0.690 }     & \textbf{0.387  }        \\
		0.15          & 1         & 0.0934          & 0.640      & 0.398          \\
		0.2           & 1         & 0.1132          & 0.569      & 0.436          \\ \bottomrule
	\end{tabular}
\end{center}
\end{table}

As is shown is Table \ref{table3}, FDRs in the fault 1 are the same for all methods, indicating that the fault 1 can be detected by all the methods. When k=0.05, 0.1, 0.15, the fault detection performance with negative feedback is superior to the one without negative feedback, which is reflected in the higher quality-relevant fault detection rate and lower thresholds of process-relevant subspace and quality-relevant subspace. When k=0.2, the fault detection performance with negative feedback is not as good as the method without negative feedback because the high-frequency oscillation occurs when the negative feedback rate is too high, which will lead to the instability of the system.

The detection performance is better when k=0.1. This indicates that a small negative feedback rate is beneficial to the fault detection performance, but an excessively large negative feedback rate will degrade the fault detection performance. Therefore, choosing an appropriate negative feedback rate can improve the fault detection performance, indicating the effectiveness of REB.

To further analyze the performance differences in detection between REB based on negative feedback and the REB based on uncertainty modeling proposed in the paper, fault detection rates of fault 1 and fault 2, thresholds of process-relevant subspace and quality-relevant subspace are shown in detail in Fig. \ref{fig10}.

In Fig. \ref{fig11}, although a small negative feedback rate can improve the detection performance, REB based on uncertainty modeling contains higher quality-relevant fault detection rate and lower thresholds of process-relevant subspace and quality-relevant subspace, showing the effectiveness of uncertainty modeling.

\subsection{Wastewater treatment process}

The urban wastewater treatment process is a long time-delay process with high coupling and complicated nonlinearity when the disturbance occurs in the wastewater treatment process \cite{Han2014,Revollar2017,Han2020}. To ensure the safe operation of the sewage treatment process and make the discharge products reach the wastewater discharge standard, it is essential to develop a monitoring method for the wastewater treatment process.

\begin{figure}[t]
	\vskip 0in
	\begin{center}
		\centerline{\includegraphics[width=\columnwidth]{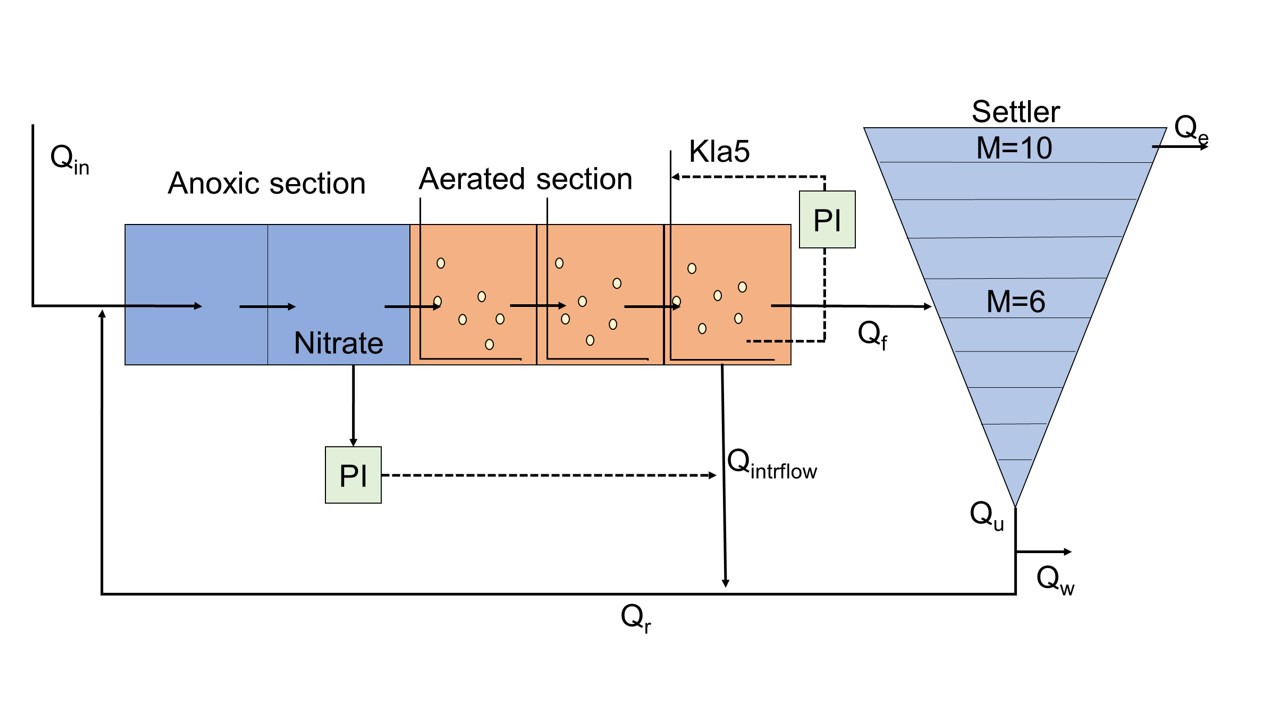}}
		\caption{The process flow chart of BSM1}
		\label{fig12}
	\end{center}
	\vskip -0.4in
\end{figure}

BSM1 is a widely used wastewater treatment process model proposed by European Co-operation in the field of Scientific and Technical Research (COST) for the institutions of water treatment research around the world. As is shown in Fig. \ref{fig12} \cite{Chen2020a}, the BSM1 is composed of a bioreactor and a clarifier. The bioreactor includes two anoxic sections and three aerated sections. It combines nitrification with pre-denitrification and is commonly used to achieve bioreactor denitrification in full-scale plants. The water in the biochemical reaction pool flows out of the fifth unit, some flows into the secondary sink from the sixth layer, and the other flows back through the pipe to the inlet of the biochemical treatment unit for internal circulation. The secondary sedimentation tank contains non-reactive units which are artificially divided into ten layers. Part of the units that conform to sewage discharge standards are discharged from the top to the river, while the sludge clarifier at the bottom is divided into two parts, one part is sent back to the first anoxic tank, and the other part is pumped out of the remaining sludge. 

COST provides three different weather conditions of the inflow water described in the files (http://www.benchmarkwwtp.org/) for simulating BSM1, namely dry, rain, and stormy weather conditions. The weather file contains 14 days of sampled data at intervals of 15 minutes. In this experiment, the BSM1 model is used to simulate the dry weather input data by 1,344 times. The first 672 samples are selected as the training dataset, and the last 672 samples are selected as the validation dataset. The descriptions of the 15 variables considered by our simulation are shown in Table \ref{table4} \cite{Hong2020}:

\begin{table}[!htb]
	\caption{The specific meanings and units of monitoring variables}
	\label{table4}
	\begin{center}
		\resizebox{\textwidth}{!}
			{
			\begin{tabular}{ccc}
				\hline
				Variable & Description & Unit\\
				\hline	
				
				${T_N}$ & The concentration of total nitrogen in the effluent & $gN  \cdot {m}^{-3}$\\
				$T_{COD}$ & The concentration of total COD in the effluent & ${g} {COD} \cdot {m}^{-3}$\\
				${S}_\mathrm{{NH}_ {e}}$ & The concentration of ${NH}_{4}^{+}+{NH}_{3}$ nitrogen in the effluent & ${g} {N} \cdot {m}^{-3}$\\
				${T}_{{SS}}$ & The total amount of solids in the effluent & ${g} {SS} \cdot {m}^{-3}$\\
				$S_{S_ i}$&The concentration of readily biodegradable substrate of influent  &${g}{COD} \cdot {m}^{-3}$\\
				$S_{NH_ i}$& The concentration of ${NH}_{4}^{+}+{NH}_{3}$ nitrogen in the influent & ${g} {N} \cdot {m}^{-3}$\\
				$S_{O_ 3}$& The concentration of dissolved oxygen in $3^{r d}$ reactor& ${g}{(-COD)} \cdot {m}^{-3}$\\
				$S_{NO_ 3}$& The concentration of ${NH}_{4}^{+}+{NH}_{3}$ in $3^{r d}$ reactor& ${g} {N} \cdot {m}^{-3}$\\
				$S_{NH_ 3}$& The concentration of ${NH}_{4}^{+}+{NH}_{3}$ in $3^{r d}$ reactor& ${g}{N} \cdot {m}^{-3}$\\
				$S_{O_ 4}$& The concentration of dissolved oxygen in $4^{t h}$ reactor& ${g}{(-COD)} \cdot {m}^{-3}$\\
				$S_{O_ 5}$& The concentration of dissolved oxygen in $5^{t h}$ reactor& ${g} {(-COD)} \cdot {m}^{-3}$\\
				$S_{NO_ 5}$& The concentration of nitrate and nitrite nitrogen in $5^{t h}$ reactor& ${g} {N} \cdot {m}^{-3}$\\
				$S_{NH_ 5}$& The concentration of ${NH}_{4}^{+}+{NH}_{3}$ in $5^{t h}$ reactor& ${g}{N} \cdot {m}^{-3}$\\
				$S_{S_ 5}$&The concentration of readily biodegradable substrate in $5^{t h}$ reactor&${g} {COD} \cdot {m}^{-3}$\\
				$BOD5$ & The concentration of total BOD5 in the effluent & ${g}{BOD} \cdot{m}^{-3}$\\
				\hline
			\end{tabular}
			}
	\end{center}
\end{table}

BOD5 is set as the quality-relevant variable, and the remaining 14 variables are set as input variables. The fault data are set from the 673rd sample with the description in table \ref{table5}.

\begin{table}[!htb]
	\caption{The specific meanings and units of monitoring variables}
	\label{table5}
	\begin{center}
		\resizebox{\textwidth}{!}
			{
			\begin{tabular}{ccc}
				\hline
				Fault & Description & Type\\
				\hline	
					
				1 & \tabincell{c}{The maximum specific growth rate of autotrophic\\ bacteria in the first aerobic basin}  & & Step \\
				2 & The oxygen transfer coefficient of the fourth bioreactor & Step \\
				3 & The measured value of $S_{NO_ 2}$ sensor in the second reactor & Step \\
				\hline
				\end{tabular}
		}
	\end{center}
\end{table}

The first fault is a step-change in the maximum specific growth rate of autotrophic bacteria in the first aerobic pool, which decreased from 0.5 to 0.3. Due to the weakened life activities of the autotrophic bacteria, the complex biochemical reactions of the wastewater treatment process are affected. The second fault is halving of the oxygen transfer coefficient in the fourth bioreactor. The change in the concentration of dissolved oxygen will change the removal performance of organic matter by the decomposition of heterotrophic bacteria and the nitrogen dislodgement of nitrification of autotrophic bacteria and denitrification of heterotrophic bacteria, and thus affect the biochemical reactions of the wastewater treatment process. The third fault case is when the $S_{NO_2}$ sensor produces a displacement range of 0.5. As the measurement of the concentration of nitrate and nitrite nitrogen in the second reactor is the input of the controller, the disagreement of the corresponding real value will affect the wastewater treatment process.

\begin{figure}[]
	\vskip 0in
	\begin{center}
		\centerline{\includegraphics[width=\columnwidth,height=20cm]{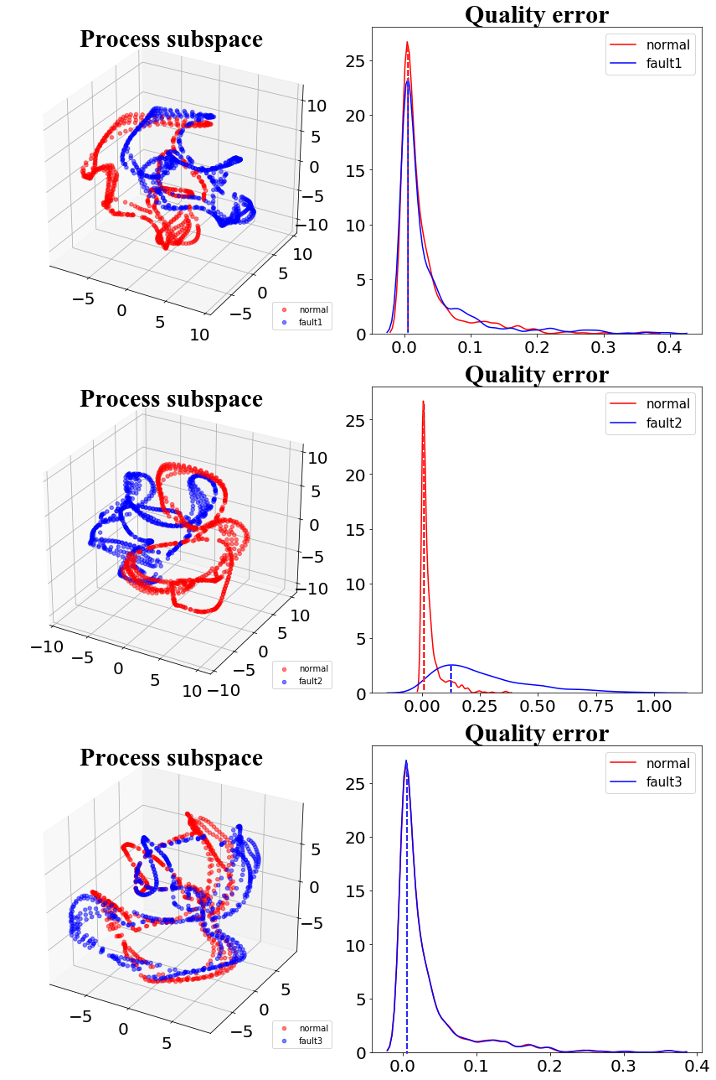}}
		\caption{Visualization of faults in BSM1}
		\label{fig13}
	\end{center}
	\vskip -0.4in
\end{figure}

Similar to the numerical example, the characteristics of the process-relevant subspace and the quality-relevant subspace are visualized in Fig. \ref{fig13}. In the process-relevant subspace, a three-dimensional scatterplot based on the t-SNE of the process variables is recorded. In the quality-relevant subspace, the KDEs of the errors between the label of the quality indicators and the estimation based on PLS are recorded. Different from the numerical example, the process-relevant subspace in BSM1 shows the correlation between samples, that is the distance between the features of adjacent samples pre-processed by t-SNE is small. This is not surprising as the continuous process show similarities in adjacent samples, which is reflected in small European distance of adjacent samples. In addition, the differences between the normal data and the abnormal data exists, in which the differences in the fault 2 is obvious. In the quality-relevant subspace, the KDEs between the normal data and the fault data of the fault 2 are obviously different, indicating that the fault 2 is an obvious quality-relevant fault. The KDEs between the normal data and the fault data of the the fault 3 are basically the same, indicating that the fault 3 is a quality-irrelevant fault. The KDEs between the normal data and the fault data of the fault 1 contain insignificant differences, which indicates that the quality indicators are influenced by fault 1 slightly. However, due to the accuracy of the modeling of PLS, the change of the quality indicators cannot be detected.

\begin{figure}[t]
	\vskip 0in
	\begin{center}
		\centerline{\includegraphics[width=\columnwidth]{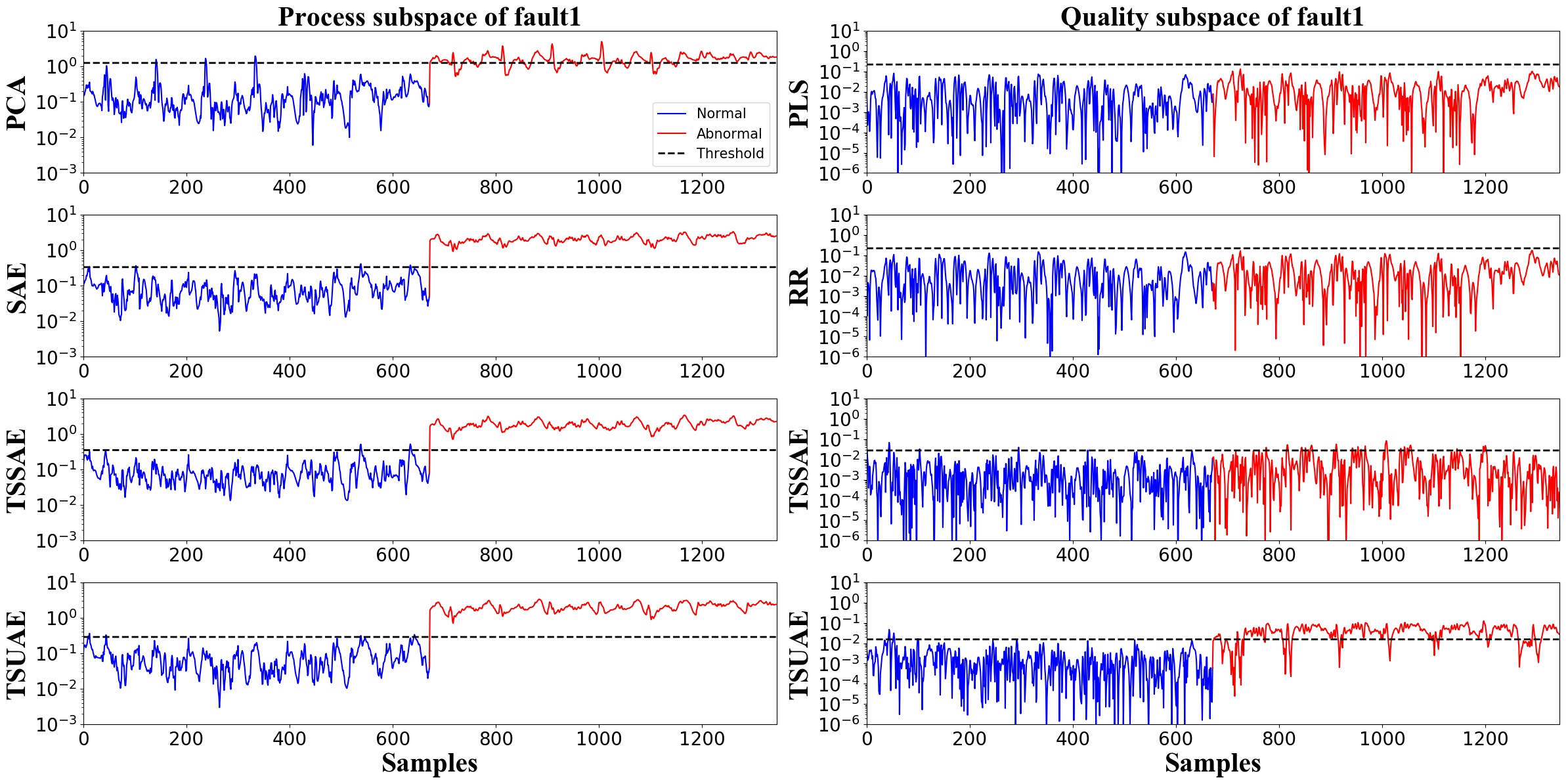}}
		\caption{The fault detection performance of fault 1 in BSM1}
		\label{fig14}
	\end{center}
	\vskip -0.4in
\end{figure}

\begin{figure}[t]
	\vskip 0in
	\begin{center}
		\centerline{\includegraphics[width=\columnwidth]{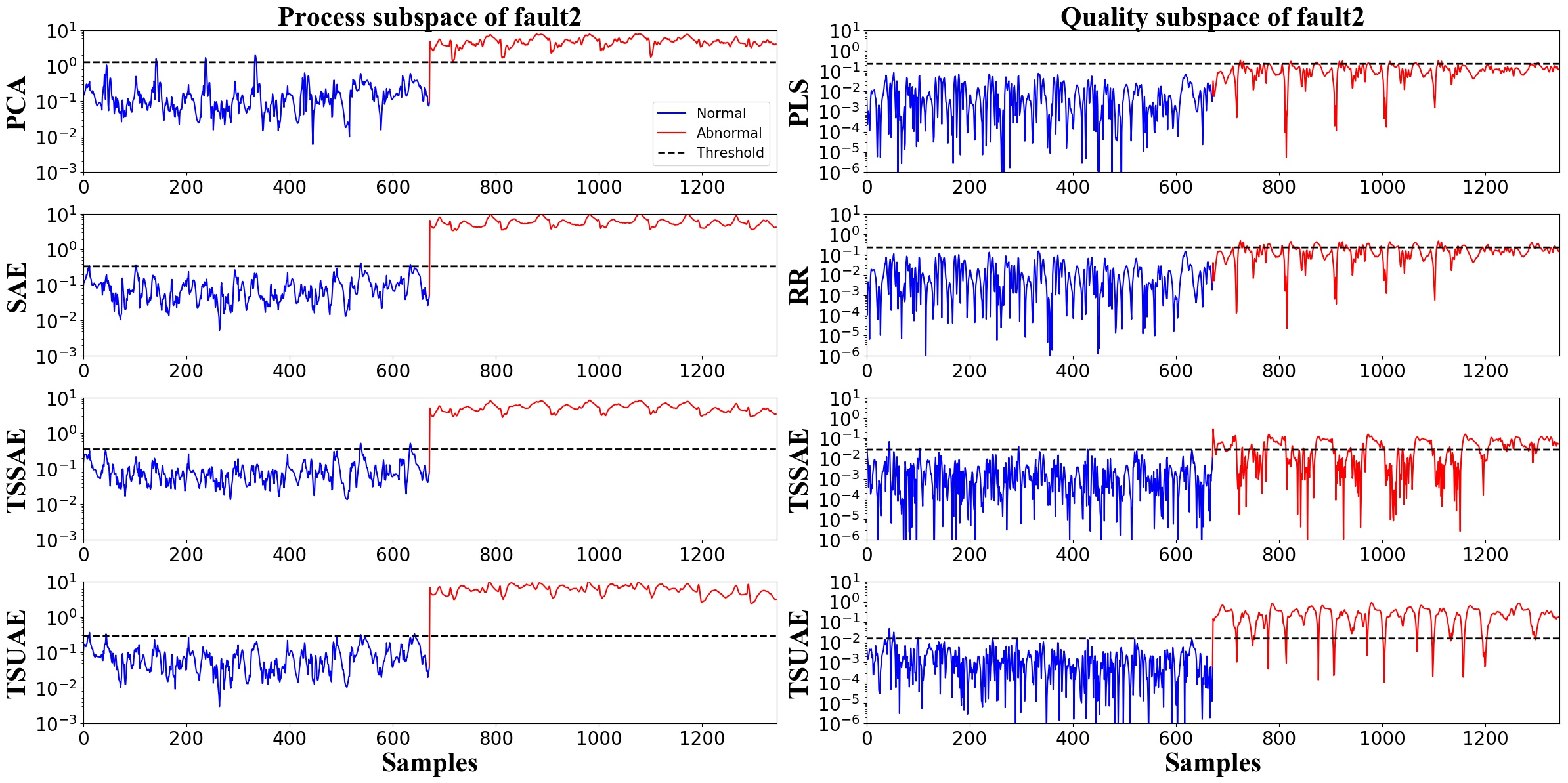}}
		\caption{The fault detection performance of fault 2 in BSM1}
		\label{fig15}
	\end{center}
	\vskip -0.4in
\end{figure}

\begin{figure}[t]
	\vskip 0in
	\begin{center}
		\centerline{\includegraphics[width=\columnwidth]{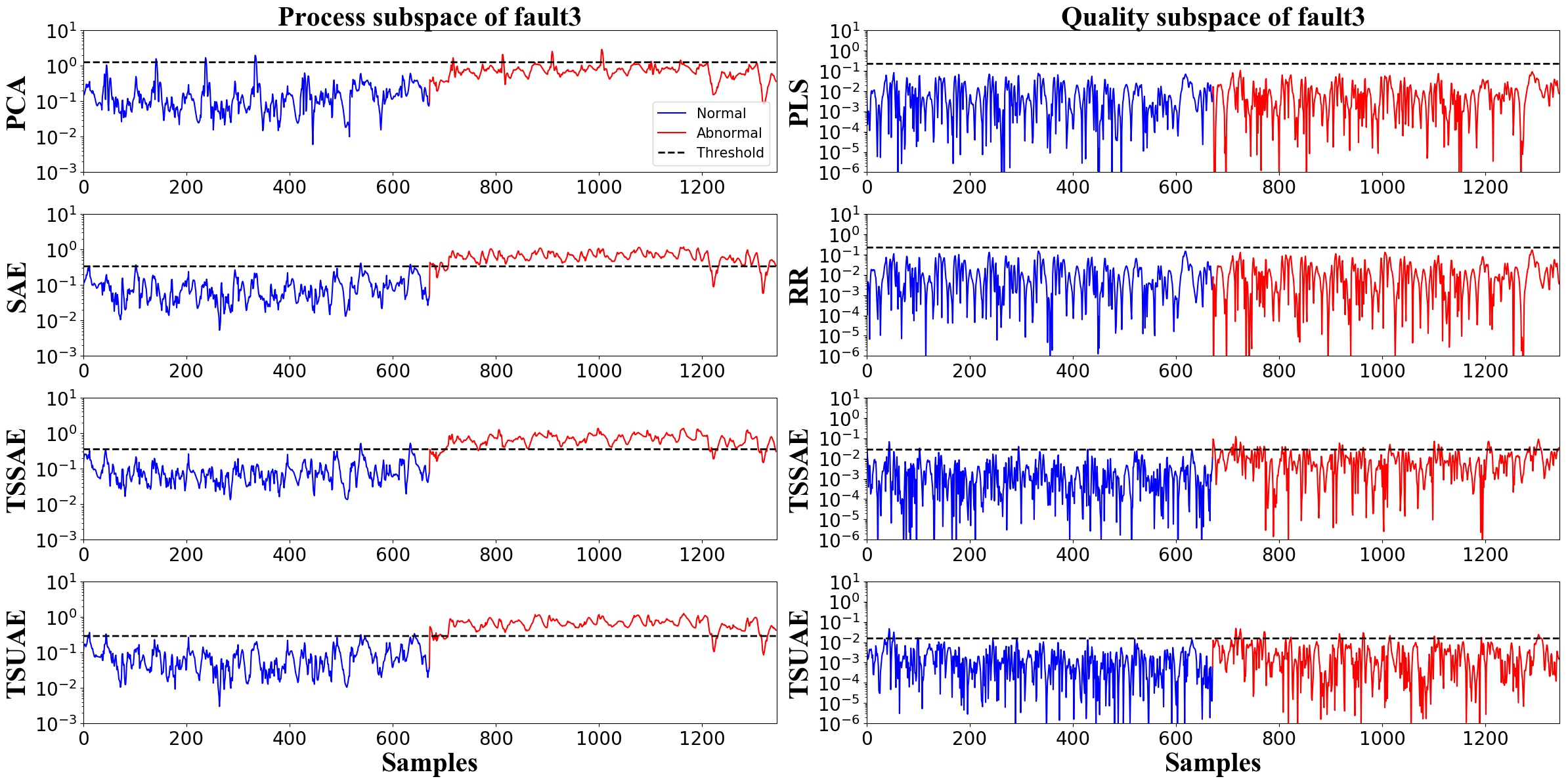}}
		\caption{The fault detection performance of fault 3 in BSM1}
		\label{fig16}
	\end{center}
	\vskip -0.4in
\end{figure}

To detect these faults and analyze the influence on the quality-relevant variables, we monitor the fault in the process-relevant subspace and analyze the fault in the quality-relevant subspace, the result of which is shown in Figs. \ref{fig14}, \ref{fig15} and \ref{fig16} through logarithmic monitoring graphs. 

\begin{table}[!htb]
	
	\caption{FARs/FDRs of three faults in process-relevant  subspace and quality-relevant  subspace}
	\label{table6}
	\begin{center}
		\resizebox{\textwidth}{!}
		{
		\begin{tabular}{@{}ccccccc@{}}
			\hline
			&&&&&& \\[-10pt] 
			\multirow{2}{*} & \multicolumn{2}{c}{1}      & \multicolumn{2}{c}{2} & \multicolumn{2}{c}{3} \\ \cmidrule(l){2-7} 
			& Dx & Dy & Dx & Dy & Dx & Dy \\ \midrule
			\textbf{TSUAE} & 0.01/\textbf{1.00} & 0.01/\textbf{0.81} & 0.01/\textbf{1.00} &0.01/ \textbf{0.94} &0.01/ \textbf{0.93} & 0.01/0.05 \\
			\textbf{TSSAE} & 0.01/\textbf{1.00} & 0.01/0.06 & 0.01/\textbf{1.00} &0.01/ 0.77 &0.01/ 0.90 & 0.01/\textbf{0.09} \\
		\textbf{SAE} & 0.01/\textbf{1.00} & / & 0.01/\textbf{1.00} & / &0.01/ 0.92 & / \\
			\textbf{PCA} & 0.01/0.77 & / & 0.01/\textbf{1.00} & / &0.01/ 0.03& / \\
			
			\textbf{PLS} & / &0.01/ 0.03 & / & 0.01/0.45 & / & 0.01/0.01 \\
			\textbf{RR} & / & 0.01/0.01 & / & 0.01/0.62 & / & 0.01/0.01 \\
			\hline       
		\end{tabular}
	}
	\end{center}
\end{table}

FARs of all methods are relatively low because the thresholds are determined by KDE, which can guarantee the reliability of FDRs. Thus, we concentrate on analyzing FDRs. In the process-relevant subspace, all the fault detection methods can detect fault 1 and fault 2, which indicating these faults show unignorable changes in process-relevant subspace. In the fault 3, the nonlinearity of the process dominates the prediction error in the statistics, and thus PCA fails to detect the fault as it substitutes the nonlinear process as the linear process. However, the detection performance of neural network-based methods SAE, TSSAE, and TSUAE is better. As a whole, our proposed method has better performance than other methods.

In the quality-relevant subspace, we analyze each fault to show the performance of the methods. SAE and PCA have no mechanisms for the detection of quality-relevant faults, so they will not be included in this part. For the fault 1, PLS, RR, and TSSAE cannot detect the fault, but our proposed TSUAE can detect the fault. However, it should be mentioned that the quality-relevant statistics of TSSAE have been changed. According to the mechanisms of the process, it can be understood that the changes in the maximum growth rate of autotrophic bacteria in the first anoxic section will affect the quality-relevant variables. Since the changes in quality-relevant variables are not large, it is considered an incipient quality-relevant fault. 

For the fault 2, all quality-relevant fault detection methods (TSSAE, TSUAE, PLS, RR)  can detect the fault. The changes of quality-relevant variables indicated the concentration of dissolved oxygen will affect the quality-relevant variables, which is considered as an obvious quality-relevant fault. Especially, our proposed method has the highest detection performance compared to other methods, which suggests that our method is more sensitive to quality-relevant failures.

Interestingly, for the fault 3, none of the methods can detect the fault. If we further analyze the fault, it can be considered that the fault 3 is caused by the sensor failure instead of the process failure. Therefore, the fault cannot be detected during the process-relevant and quality-relevant subspaces. Thus, the fault is considered as a quality-irrelevant fault.

In summary, compared to other methods, our proposed method can more efficiently detect not only the largely quality-relevant faults but also the incipient quality-relevant fault in most situations. It can be concluded that TSUAE achieves better detection results, compered to TSSAE, PLS and RR.

Finally, as the lower threshold value of detection methods under the same statistics represents more meaningful representative features of data, the thresholds of different methods are recorded, as shown in Table \ref{table7}. It is obvious TSUAE has a lower threshold value in the process-relevant subspace and the quality-relevant subspace compared to other methods. 

\begin{table}[!htb]
	\caption{The thresholds of different methods in the BSM1}
	\label{table7}
	\begin{center}
		\begin{tabular}{@{}ccccccc@{}}
			\hline
			&&&& \\[-10pt] 
		 & TSSAE & TSUAE & SAE & PCA & PLS & RR \\ 
		 \hline
			Dx &\textbf{0.290} & 0.353 & 0.334 & 1.256 & / & / \\ 
			Dy & \textbf{0.015}& 0.028 & / & / & 0.222 & 0.115 \\
			\hline       
		\end{tabular}
	\end{center}
\end{table}

\section{Conclusion }

In the traditional teacher-student network, the performance of the student network is inferior to that of the teacher network, which is caused by the features differences between the student network and the teacher network. Thus, in this paper, REB, a novel block evaluates the differences between teacher network and student network and reduces the performance degeration of the student network, is proposed. To generate more realistic feedback than transferring the features differences scaled by a pre-set negative feedback rate and reduce the impact caused by differences fluctuations, uncertainty modeling is introduced to estimate the differences from REB. Accordingly TSUAE is proposed. In TSUAE, the teacher network and the student network are trained iteratively at the same time according to the feedback obtained from REB, which is beneficial to reduce the difference between the teacher network and the student network. Then, a quality-relevant fault detection method based on the TSUAE is proposed. The recovery of the evaluated process inputs and the prediction of the quality-relevant indexes are calculated by the teacher and student framework so that we can effectively monitor faults both in the process-relevant subspace and quality-relevant subspace. The fault detection method is proved to be effective by a numerical simulation and a sewage treatment process simulation.

However, the characteristics of input data of the teacher network and the student network are considered to be similar in this paper, but actually they maybe inconsistent, which is not considered in the proposed method. As REB and uncertainty modeling can measure the performance degradation through modeling, how to apply the proposed method to the situation where the characteristics of input data of the training data and test data is our future work.

\bibliographystyle{IEEEtran}
\bibliography{TSUAE}







\end{document}